\documentclass[letterpaper,10pt,journal,twoside]{IEEEtran}

\usepackage{graphics} 
\usepackage{bm}
\usepackage{epsfig} 
\usepackage{times} 
\usepackage{amsmath} 
\usepackage{amssymb}  
\usepackage{amsfonts} 

\usepackage{newclude} 

\usepackage[textsize=footnotesize]{todonotes}

\usepackage{graphicx}

\usepackage[ruled]{algorithm2e}

\usepackage{amsthm}
\usepackage{amsfonts}
\usepackage{mathtools}
\usepackage{enumitem}
\usepackage[group-separator={,}]{siunitx} 

\usepackage{tabularx,colortbl}
\usepackage{booktabs}

\usepackage[outdir=export/]{epstopdf}

\usepackage{multirow}
\usepackage{wrapfig}

\usepackage{algorithmic}
\usepackage{array}
\newcounter{rowcntr}[table]
\usepackage[font={small,it},belowskip=0pt]{caption}
\usepackage{placeins} 

\usepackage[normalem]{ulem}

\usepackage[percent]{overpic}
\usepackage{hyphenat}

\usepackage{tabu}
\usepackage{xcolor}
\usepackage{textcomp}

\usepackage[nolist,nohyperlinks]{acronym}
\usepackage{mathtools}

\usepackage{adjustbox}
\usepackage{float}
\usepackage{xspace}
\usepackage{empheq} 

\graphicspath{{drawings/}{export/}}

\newcommand{\vc}[1]{\ensuremath{\bm{#1}}}

\newcommand{\bmat}[1]{\ensuremath{\begin{bmatrix}#1\end{bmatrix}}}

\DeclarePairedDelimiter\abs{\lvert}{\rvert}
\DeclarePairedDelimiter\norm{\lVert}{\rVert}
\makeatletter
\let\oldabs\abs
\let\oldnorm\norm
\def\abs{\@ifstar{\oldabs}{\oldabs*}}
\def\norm{\@ifstar{\oldnorm}{\oldnorm*}}
\makeatother

\DeclareMathOperator{\rank}{rank}

\DeclareMathOperator{\ve}{vec}

\DeclareMathOperator{\tr}{tr}

\def\-{\raisebox{0pt}{-}}

\newcommand{\vech}[1]{\ensuremath{\operatorname{vech}\!\big( #1 \big)}}

\newcommand{\nullsp}[1]{\ensuremath{\operatorname{null}(#1)}}
\newcommand{\mataug}[1]{\ensuremath{\operatorname{mat}\!\big( #1 \big)}}
\newcommand{\vecaug}[1]{\ensuremath{\operatorname{vech}\!\big( #1 \big)}}

\newcommand{\vechinv}[1]{\ensuremath{\operatorname{vech}^{-1}\!\big( #1 \big)}}

\def\vectorised#1{\ve\left(#1\right)}

\newcommand{\R}{\ensuremath{\boldsymbol{\mathbb{R}}}}
\newcommand{\inR}[1]{\ensuremath{\in \R^{#1}}}
\newcommand{\inS}[1]{\ensuremath{\in \boldsymbol{\mathbb{S}}_+^{#1}}}

\newcommand{\SO}[1]{\ensuremath{SO({#1})}}
\newcommand{\Orth}[1]{\ensuremath{O({#1})}}
\newcommand{\SE}[1]{\ensuremath{SE({#1})}}
\renewcommand{\det}[1]{\text{det}\left(#1\right)}
\newcommand\kron{\ensuremath{\otimes}}

\newcommand{\inner}[1]{\ensuremath{\langle #1 \rangle}}
\newcommand{\inindex}[1]{\ensuremath{\in[#1]}}
\newcommand{\barr}{\,\big|\,}

\newtheoremstyle{example}%
{}
{}
{\itshape}
{}
{\bfseries}
{\normalfont{.}}
{ }
{}
\newtheoremstyle{definition}%
{}
{}
{\itshape}
{}
{\bfseries}
{\normalfont{.}}
{ }
{}
\newtheoremstyle{theorem}%
{}
{}
{\itshape}
{}
{\bfseries}
{\normalfont{.}}
{ }
{}
\theoremstyle{definition}
\newtheorem*{definition}{Definition}
\theoremstyle{example}
\newtheorem*{example}{Example}
\theoremstyle{theorem}
\newtheorem{theorem}{Theorem}
\theoremstyle{definition}
\newtheorem{lemma}{Lemma}

\let\oldunderbrace\underbrace
\renewcommand{\underbrace}[2]{\let\scriptstyle\textstyle \oldunderbrace{#1}_{#2}}
\definecolor{tab10blue}{rgb}{0.12156863, 0.46666667, 0.70588235}

\def\vs{\emph{vs}.}
\def\eg{\emph{e.g}.} 
\def\ie{\emph{i.e}.} 
 
\def\etc{\emph{etc}.} 
 
\def\wrt{w.r.t.}
\def\etal{\emph{et al.~}}

\DeclareFontFamily{U}{mathx}{\hyphenchar\font45}
\DeclareFontShape{U}{mathx}{m}{n}{ <-> mathx10 }{}
\DeclareSymbolFont{mathx}{U}{mathx}{m}{n}
\DeclareFontSubstitution{U}{mathx}{m}{n}
\DeclareMathAccent{\widebar}{\mathalpha}{mathx}{"73}
\newcolumntype{C}[1]{>{\centering\let\newline\\\arraybackslash\hspace{0pt}}m{#1}}
\newcolumntype{D}[1]{>{\centering\let\newline\\\arraybackslash\hspace{0pt}\columncolor{gray!20}}m{#1}}
\newcolumntype{R}[1]{>{\raggedleft\let\newline\\\arraybackslash\hspace{0pt}}m{#1}}
\newcolumntype{L}[1]{>{\raggedright\let\newline\\\arraybackslash\hspace{0pt}}m{#1}}
\newcolumntype{M}[1]{>{\raggedright\let\newline\\\arraybackslash\hspace{0pt}}m{#1}}
\newcolumntype{N}{>{\refstepcounter{rowcntr}\therowcntr}c}

\usepackage{tabstackengine}
\stackMath
\setstackgap{L}{18pt}
\setstacktabbedgap{0pt}
\def\lrgap{\kern6pt}

\newcommand{\hc}[1]{\ensuremath{\hat{\vc{#1}}}}
\newcommand{\oc}[1]{\ensuremath{{\vc{#1}^\star}}}

  \renewenvironment{thebibliography}[1]{%
    \begin{oldthebibliography}{#1}%
      \setlength{\parskip}{0ex}%
      \setlength{\itemsep}{0ex}%
  }%
  {%
    \end{oldthebibliography}%
  }

\def\autotight{\textsc{AutoTight}\xspace}
\def\autoscale{\textsc{AutoTemplate}\xspace}
\def\autotemplate{\textsc{AutoTemplate}\xspace}

\def\secondmethod{\autoscale\xspace}
\def\ppr{PPR\xspace}
\def\plr{PLR\xspace}
\def\rppr{rPPR\xspace}
\def\rplr{rPLR\xspace}

\def\hom{\ensuremath{h}}
\DeclareMathSymbol{\shortminus}{\mathbin}{AMSa}{"39}
\newcommand{\fub}{\ensuremath{b_u}}
\let\mc\multicolumn

\acrodef{QCQP}[QCQP]{quadratically constrained quadratic program}
\acrodef{SDP}[SDP]{semidefinite program}
\acrodef{EVD}[EVD]{eigenvalue decomposition}
\acrodef{SVD}[SVD]{singular value decomposition}
\acrodef{MAP}[MAP]{maximum a posteriori}
\acrodef{GP}[GP]{Gaussian process}
\acrodef{LTV}[LTV]{linear, time-varying}
\acrodef{SLAM}[SLAM]{simultaneous localization and mapping}
\acrodef{IMU}[IMU]{inertial measurement unit}
\acrodef{psd}[PSD]{positive-semidefinite}
\acrodef{GN}[GN]{Gauss-Newton}
\acrodef{UWB}[UWB]{ultra-wideband}
\acrodef{rmse}[RMSE]{root-mean-squared error}
\acrodef{mae}[MAE]{mean absolute error}
\acrodef{LM}[LM]{Levenberg-Marquardt}
\acrodef{GPS}[GPS]{global positioning system}
\acrodef{AUV}[AUV]{autonomous underwater vehicle}
\acrodef{SOS}[SOS]{sums-of-squares}
\acrodef{RO}[RO]{range-only}
\acrodef{KKT}[KKT]{Karush-Kuhn-Tucker}
\acrodef{TLS}[TLS]{truncated least-sqares}
\acrodef{ER}[ER]{eigenvalue ratio}
\acrodef{POP}[POP]{polynomial optimization problem}
\acrodef{RDG}[RDG]{relative duality gap}
\acrodef{AE}[AE]{associate editor}
\acrodef{NLS}[NLS]{nonlinear least-squares}
\acrodef{STD}[STD]{standard deviation}

\usepackage[switch]{lineno} 
\usepackage{titlesec}

\newcommand{\last}[1]{\textcolor{red!70}{#1}}
\newcommand{\pol}[1]{\textcolor{blue!70}{#1}}

\renewcommand\last[1]{#1}
\renewcommand\pol[1]{#1}
\renewcommand\todo[1]{}

\newcommand\myheader{\framebox{This paper was accepted to IEEE Transactions on Robotics, 2024 (\href{https://doi.org/10.1109/TRO.2024.3454570}{link}).}}
\usepackage{fancyhdr}
\fancypagestyle{firstpage}
{
  \fancyhead[L]{ \centering \myheader}
  \fancyhead[R]{}
}
\fancypagestyle{empty}
{
  \fancyhead[L]{}
  \fancyhead[R]{}
}

\title{Toward Globally Optimal State Estimation Using Automatically Tightened Semidefinite Relaxations} 
\author{Frederike Dümbgen~\IEEEmembership{Member, IEEE}, Connor Holmes~\IEEEmembership{Student Member, IEEE}, Ben Agro, Timothy~D.~Barfoot~\IEEEmembership{Fellow, IEEE}
  \thanks{FD is with Inria, École Normale Supérieure, PSL University, Paris, France. CH, BA and TDB are with the University of Toronto Robotics Institute, University of Toronto, Canada. Corresponding author: {\tt\footnotesize frederike.dumbgen@inria.fr}}
  \thanks{Received 3 April 2024; accepted 19 July 2024. Date of publication 4 September 2024;}
  \thanks{The majority of this work was conducted while FD was at University of Toronto. The work of FD was supported in part by the ANR JCJC through Project NIMBLE under Grant ANR-22-CE33-0008 and in part by the SNF, Switzerland, Postdoc Mobility Grant under Grant 206954. The work of CH and TDB was supported by NSERC, Canada.}
}
\usepackage{hyperref}  
\hypersetup{colorlinks=true,linkcolor=gray,urlcolor=gray,citecolor=gray}

\begin{document} 

\maketitle 
\renewcommand{\headrulewidth}{0pt}
\thispagestyle{firstpage}

\begin{abstract}
In recent years, semidefinite relaxations of common optimization problems in robotics have attracted growing attention due to their ability to provide globally optimal solutions. In many cases, it was shown that specific handcrafted redundant constraints are required to obtain tight relaxations and thus global optimality. These constraints are formulation-dependent and typically identified through a lengthy manual process. Instead, the present paper suggests an automatic method to find a set of sufficient redundant constraints to obtain tightness, if they exist. We first propose an efficient feasibility check to determine if a given set of variables can lead to a tight formulation. Secondly, we show how to scale the method to problems of bigger size. At no point of the process do we have to find redundant constraints manually. We showcase the effectiveness of the approach, in simulation and on real datasets, for range-based localization and stereo-based pose estimation. \pol{We also} reproduce semidefinite relaxations presented in recent literature and show that our automatic method always finds a smaller set of constraints sufficient for tightness than previously considered.

\begin{IEEEkeywords}
Optimization and optimal control, Localization, Robot Safety, Global Optimality
\end{IEEEkeywords}
\end{abstract}

\section{Introduction}

\IEEEPARstart{M}{any} problems encountered in robotic state estimation, such as calibration and \ac{SLAM}, are typically posed as~\ac{NLS} optimization problems~\cite{rehder_extending_2016,barfoot_state_2017}. Widely adopted solvers for these problems, such as \ac{GN} and \ac{LM}, have only local, if any, convergence guarantees and may return suboptimal solutions~\cite{nocedal_numerical_2006}.

Over the past years, there has been a growing effort to exploit semidefinite relaxations of these optimization problems. \pol{They} open the door to global optimality in at least two different ways: in certain cases, a (convex)~\ac{SDP} (or a sequence thereof) may be solved instead of the original nonconvex problem to find the globally optimal solution~\cite{papalia_certifiably_2023, rosen_se-sync_2019,yang_inexact_2021,yang_certifiably_2022}. 
In other cases, the Lagrangian dual of the \ac{SDP} offers the possibility to construct so-called \emph{optimality certificates}~\cite{yang_teaser_2020,eriksson_rotation_2018} to determine the global optimality of the solutions obtained by local solvers.

The performance and feasibility of the aforementioned methods greatly depends on whether the \ac{SDP} relaxation is \pol{\emph{tight}}. For example, for some problems, the globally optimal solution to the original problem can only be extracted from the~\ac{SDP} solution when \pol{the latter} has rank one, \pol{in which case the relaxation is called tight}~\cite{yang_teaser_2020,briales_convex_2017}. Similarly, \pol{common} certifiable algorithms work only when strong duality \pol{holds}~\cite{boyd_convex_2004},~\ie, when the cost of the relaxed problem equals the cost of the original problem\pol{, which is also sometimes referred to as tightness}~\cite{yang_teaser_2020,rosen_se-sync_2019}.\footnote{\pol{To disambiguate between these two cases, we will use the terms \emph{rank tightness} and \emph{cost tightness} in this paper.}}

\begin{figure}[t]
  \centering
  \includegraphics[width=\linewidth]{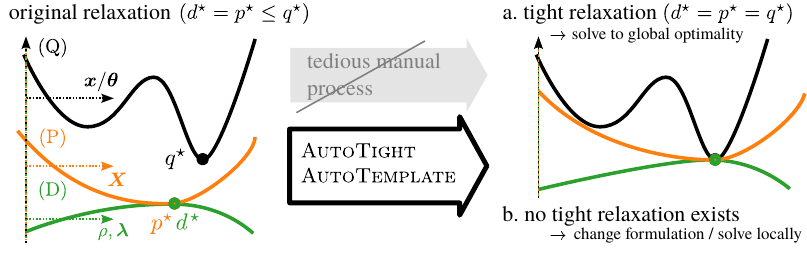}
  \caption{The proposed method in a nutshell: we circumvent the lengthy manual process of finding redundant constraints to tighten a given semidefinite relaxation, using instead a sampling-based approach to automatically find \pol{them}. This allows for the quick evaluation of different formulations and substitutions of a given optimization problem, \pol{enabling} \ac{SDP}s to be more widely adopted for finding globally optimal solutions to \pol{state estimation} problems in robotics.}
  \label{figure1}
\end{figure}

One important enabler for tight relaxations has been a mathematical framework called \textit{Lasserre's hierarchy}~\cite{lasserre}. Put simply, the hierarchy consists of a sequence of semidefinite relaxations where polynomial terms of increasing order are added to the original problem. Calling the original variable dimension $d$ and the hierarchy order $k$, each level \pol{consists of} a $N_k$-dimensional~\ac{SDP}, with $N_k:=\tbinom{d+k}{k}$. Under weak technical assumptions, any problem that can be written as a~\ac{POP} can be lifted to a high enough order $k$ to allow for a tight relaxation. The required order may be infinite, but many follow-up works have shown that tightness is obtained with finite $k$~\cite{ruiz_using_2011,yang_teaser_2020,majumdar_convex_2014,sun_certifiably_2020}. More recently, it has been shown that many problems admit a \textit{sparse Lasserre's hierarchy}, meaning that only some of the $N_k$ terms may be required at each level~\cite{sparse-Lasserre,yang_certifiably_2022}.

As~\ac{SDP}s scale poorly with problem dimension, it is desirable to achieve tightness with as few additional higher-order substitutions as possible (ideally, with none). For this matter, it has been shown that so-called redundant constraints are paramount~\cite{yang_teaser_2020,briales_convex_2017,ruiz_using_2011,briales_certifiably_2018}. However, to date, these constraints are usually the result of a lengthy manual search process and it is often hard to retrace how the constraints were discovered. In~\cite{yang_certifiably_2022}, a method to find all ``trivially satisfied'' constraints is provided, but this process is costly and not all of these constraints may be necessary. Futhermore, using different formulations may lead to entirely different forms and numbers of required redundant constraints. Due to the lack of a systematic method of finding the right formulation and sufficient redundant constraints, practitioners often have to spend great effort in trial-and-error reformulations. This adds significant overhead as opposed to easy-to-use local solvers, and thus may hinder the wide adoption of~\ac{SDP} methods in robotics.

In this paper, we provide tools that help automate the search for redundant constraints required for tightness. We study in particular \pol{two} classical state estimation problems\pol{:} \ac{RO} localization and stereo\pol{-}camera localization, which are nonconvex and exhibit local minima in which standard solvers may get stuck~\cite{dumbgen_safe_2023}. The proposed methods enable the globally optimal solution or certification of these and many other problems presented in Section~\ref{sec:results}. More concretely, we \pol{present} two methods:
\begin{enumerate}
  \item \autotight determines  if a problem in a given form can be tightened by adding enough \pol{automatically found} redundant constraints. 
  \item \autoscale \pol{generates a set of constraint} \emph{templates} that can be \pol{applied to new problems of any numbers of variables}. 
\end{enumerate}
The focus of \autotight~is feasibility. It is purposefully kept simple \pol{(typically only a few lines of code)} and should be performed on a small example problem. The focus of \autoscale~is scalability, enabling \pol{the automatic tightening of} problems of any size, which is a \pol{strict} requirement for typically high-dimensional problems encountered in robotics. The only prerequisite for using the provided tools is a method \pol{for} randomly generating many problem setups (also called a ``sampling oracle'' in the literature~\cite{cifuentes_sampling_2017}). \pol{For many use cases, creating such a method is part of the standard development process, and if not, it can usually be obtained fairly easily. Implementations of \autotight~and \autotemplate are publicly available.}\footnote{\pol{The code is available as an open-source \texttt{Python} package at~\url{https://github.com/utiasASRL/constraint_learning}.}}

This paper is structured as follows. We put the proposed method in context with related work in Section~\ref{sec:related-work}. Then, we introduce mathematical preliminaries for relaxing a \ac{NLS} problem to an~\ac{SDP} in Section~\ref{sec:preliminaries}. In Section~\ref{sec:tightness}, we present \autotight and in Section~\ref{sec:scalability} we propose \autoscale, its scalable extension. We define the two example state-estimation problems in Section~\ref{sec:applications}, and we provide novel insights on the tightness of their relaxations, as well as other problems, in Section~\ref{sec:results}. Finally, we test the method on real-world datasets for the example applications in Section~\ref{sec:results-real} and conclude in Section~\ref{sec:discussion}.

\section{Related Work}\label{sec:related-work}

The list of problems in robotics and computer vision that have been solved using semidefinite relaxations is long and continues to grow. In vision-based state estimation, semidefinite relaxations have been widely explored, for example to solve rotation \pol{synchronization}~\cite{olsson_branch-and-bound_2009,hartley_rotation_2013,brynte_tightness_2022} or to perform camera pose estimation from pixel measurements~\cite{briales_convex_2017,briales_certifiably_2018}. The first theoretical guarantees on tightness of these and other problems were given in~\cite{cifuentes_convex_2021,eriksson_rotation_2018}. A set of analytical redundant constraints that successfully tightens many problem instances involving rotations has been proposed in~\cite{anstreicher_lagrangian_2000,briales_convex_2017} and used successfully in follow-up works to certify, for instance, hand-eye calibration~\cite{wise_certifiably_2020} and generalized-essential-matrix estimation~\cite{zhao_certifiably_2020}. Follow-up works have shown that tight relaxations can be achieved for robust cost functions, too~\cite{sun_certifiably_2020,yang_teaser_2020}, of which a \pol{comprehensive} overview, and a recipe for constructing trivially satisfied redundant constraints, is given in~\cite{yang_certifiably_2022}. Robotics planning and control problems have recently also seen a surge of relaxation-based methods~\cite{majumdar_convex_2014,marcucci_shortest_2022,marcucci_shortest_2022,graesdal_towards_2024}. Notably, specific redundant constraints (again, analytically specified) were found to be paramount for tightness in~\cite{marcucci_shortest_2022}.

For some problems, no redundant constraints are required for tightness. For these problems, methods based on the \textit{Burer Monteiro} approach~\cite{burer_local_2005} and the \textit{Riemannian staircase}~\cite{boumal_riemannian_2016} have been shown to be very effective at finding the optimal solution with speeds competitive with efficient local solvers~\cite{papalia_certifiably_2023,rosen_se-sync_2019,doherty_performance_2022,dellaert_shonan_2020}. Other methods have explored fast global optimality certificates of solutions of local solvers~\cite{holmes_efficient_2023,dumbgen_safe_2023}. To date, whenever redundant constraints are required for tightness, \ac{SDP}~solvers are generally too slow for real-time performance~\cite{yang_certifiably_2022}. However, recent advances have shown that solvers can be significantly sped up when the optimal solution is of low rank~\cite{parra_rotation_2021,yang_inexact_2021,yang_certifiably_2022}. More progress in developing fast~\ac{SDP} solvers is a requirement to enable their large-scale adoption for robotics; another requirement is finding the necessary redundant constraints for a larger class of problems. The method proposed in this paper contributes to the latter \pol{requirement}.

Recently, a sampling paradigm has been explored in the~\ac{SOS} literature to overcome some limitations of~\ac{SDP} solvers~\cite{cifuentes_sampling_2017}.\footnote{There is a \pol{direct} connection between the~\ac{SOS} relaxation and Lasserre's hierarchy (also called moment relaxation in this context); a clear description of this connection is given in~\cite{brynte_tightness_2022}.} The authors suggest to solve an~\ac{SDP} based on only a small number of feasible samples of the corresponding~\ac{SOS} program. The method thus exploits the geometry of the variety without the use of advanced concepts such as Gröbner bases~\cite{Groebner}. This solution has shown great promise on small problems in control~\cite{shen_sampling_2020}. We use a similar paradigm in this paper, but instead of solving a sampling-based \ac{SDP}, we use the samples to find generalizable constraints. \pol{This allows us to generate tight \ac{SDP} relaxations} for a wide range of problems, \pol{and by learning templates, allows us to generalize to new and higher-dimensional problems.}

The present paper complements our prior and parallel work~\cite{holmes_semidefinite_2024,dumbgen_safe_2023,goudar_optimal_2024,barfoot_certifiably_2024} as follows. We show in \cite{holmes_semidefinite_2024} that the semidefinite relaxation commonly used in pose graph optimization and~\ac{SLAM}~\cite{holmes_efficient_2023,rosen_se-sync_2019} requires redundant constraints when using non-isotropic measurement-noise models. \pol{A preliminary form of the presented methods} were used to automatically find these redundant constraints. The optimality certificate derived in~\cite{dumbgen_safe_2023} for \ac{RO} localization does not require redundant constraints, but we show in the present paper that a different formulation does, and we find these constraints automatically. For \ac{RO} pose estimation~\cite{goudar_optimal_2024} and rotation estimation with the Cayley map~\cite{barfoot_certifiably_2024}, \pol{the presented methods (in simplified form) were also used to find the redundant constraints required for tightness.}

\section{Preliminaries}\label{sec:preliminaries}

\subsection{Notation}

We denote vectors and matrices by bold-face lowercase and uppercase letters, respectively.   
The transpose of matrix \vc{A} is written as $\vc{A}^\top$. The identity matrix in $d$ dimensions is $\vc{I}_d$\pol{.}  A~\ac{psd} matrix is written as $\vc{X}\succeq 0$, and we denote the space of $N\times N$ \ac{psd} matrices by $\mathbb{S}_+^{N}$.
The operator $\otimes$ is the Kronecker product and the operator $\left\lceil{\cdot}\right\rceil$ is the ceiling function. The inner product is denoted by $\inner{\cdot,\cdot}$, and the matrix inner product is defined as $\inner{\vc{A}, \vc{B}}=\tr{(\vc{A}^\top\vc{B})}$ where $\tr{(\cdot)}$ is the trace operator.
We introduce $\operatorname{vech}(\cdot)$, which extracts the elements of the upper-triangular part of a matrix, and multiplies the off-diagonal elements by $\sqrt{2}$. This ensures that $\inner{\vc{A},\vc{B}}=\vech{\vc{A}}^\top\vech{\vc{B}}$, and is commonly used in \ac{SDP} solvers~\cite{mosek}. We denote the inverse operation by $\operatorname{vech}^{-1}(\cdot)$.
$\vc{x}[k]$ denotes the $k$-th element of vector \vc{x}, starting at $1$.
For shorter notation, we use $[N]$ for the index set $\{1, \ldots, N\}$.

\subsection{Semidefinite Relaxations}

In the remainder of this section, we provide theoretical background on semidefinite relaxations and duality theory necessary to understand this paper for the nonexpert reader. For an in-depth introduction to these topics we refer to~\cite{boyd_convex_2004,nocedal_numerical_2006}.

Most generally speaking, the subject of this paper is optimization problems of the form
\begin{equation}
  \min_{\vc{\theta}\inR{d}} \{ c(\vc{\theta}) \barr e_j(\vc{\theta})=0, j\inindex{N_e}\},
  \label{eq:original}
\end{equation}
where $\vc{\theta}$ \pol{contains the variables}, $c(\cdot)$ is the cost, and $e_i(\cdot)$ are equality constraints.\footnote{We focus on equality constraints here for the sake of clarity. Note that inequality constraints can be added as long as they can also be written as quadratic constraints in the lifted vector and thus carried forward as quadratic inequality constraints in the relaxations. We include one example of inequality constraints in Section~\ref{sec:results-other}.} In robotics, the cost is most commonly a (robust) \ac{NLS} cost function, and the constraints may enforce the \pol{variables} to lie, for example, in \SO{3} or \SE{3}~\cite{barfoot_state_2017}. The following is a simple (unconstrained) \ac{NLS} problem that we will use throughout this paper to demonstrate the theoretical concepts. More realistic cost functions, which are also of the form~\eqref{eq:original}, can be found in Section~\ref{sec:applications}.

\begin{example}[stereo-1D, NLS] Inspired by camera localization, we propose the following pedagogical example problem:
\begin{equation}
  \min_{\theta} \sum_{i=1}^N \left(u_i -\frac{1}{(\theta - m_i)}\right)^2 =: c(\theta),
  \label{eq:ex-original1}
\end{equation}
\noindent where $\theta\inR{}$ is the decision variable, $u_i\inR{}$ are measurements and $m_i\inR{}$ are known landmarks. The problem is of the form~\eqref{eq:original}, with $N_e=0$, $d=1$, and $c(\theta)$ a~\ac{NLS} function.
\end{example}

The problems in which we are interested can be lifted to a \ac{QCQP}, which is true for any polynomial optimization problem. In other words, we assume that we can rewrite~\eqref{eq:original} as
\begin{equation}
  \min_{\vc{x}\inR{N}} \{ f(\vc{x}) \barr g_j(\vc{x})=0, l_i(\vc{x})=b_i, j\inindex{N_e}, i\inindex{N_l}\cup\{0\}\},
  \label{eq:qcqp-normal}
\end{equation}
where $f$, $g_j$, and $l_i$ are quadratic in the lifted vector \vc{x}, and $b_0=1,b_{i}=0,i\inindex{N_l}$. The lifted vector is given by
\begin{equation}
  \vc{x}^\top=\bmat{\hom & \vc{\theta}^\top & z_1 & \cdots & z_{N_l}},
  \label{eq:x}
\end{equation}
where we have introduced $z_i := \ell_i(\vc{\theta})$, higher-order lifting functions of $\vc{\theta}$. By choosing enough of these substitutions, we can enforce that each substitution can itself be written as a quadratic constraint: $l_i(\vc{x})=0$.
We have added $\hom$ in~\eqref{eq:x} as a homogenization variable, enforced by $l_0(\vc{x})={\vc{x}[1]}^2=1$, which is common practice for allowing constant and linear functions to be written as quadratic functions (see, \eg,~\cite{cifuentes_local_2022}).\footnote{Technically, the first element of $\vc{x}$ may thus take the value $-1$, but this does not pose a problem as the whole vector can then be simply negated.}  We illustrate these concepts in our pedagogical example in what follows, and refer the reader to Section~\ref{sec:applications} for examples for writing more complex problems as~\ac{QCQP}s.

\begin{example}[stereo-1D, QCQP] Using the lifted vector
  \begin{equation}\label{eq:ex-lifting}
    \vc{x}^\top = \bmat{\hom & \theta & z_1 & \cdots & z_N}, \quad z_i = \ell_i(\theta):=\frac{1}{\theta - m_i},
  \end{equation}
we can rewrite~\eqref{eq:ex-original1} in the form~\eqref{eq:qcqp-normal}, 
with $f(\vc{x})=\sum_{i=1}^N {\left(\vc{x}[1]u_i - \vc{x}[2+i]\right)}^2$, and $l_i(\vc{x}) = z_i(\theta - m_i) = \vc{x}[2+i]\vc{x}[2] - \vc{x}[2+i]\vc{x}[1]m_i, i\inindex{N}$, which are all quadratic functions in~$\vc{x}$.
\end{example}

Since all functions in~\eqref{eq:qcqp-normal} are quadratic in $\vc{x}$, we can write 
\begin{equation}
  \min_{\vc{x}\inR{N}} \{ \vc{x}^\top\!\vc{Q}\vc{x} \barr \vc{x}^\top\!\vc{A}_0\vc{x}=1, \vc{x}^\top\!\vc{A}_i\vc{x}=0, i\inindex{N_A}\},
  \tag{Q}
  \label{eq:qcqp-matrix}
\end{equation}
where $\vc{Q}$ and $\vc{A}_i, i\inindex{N_A}$ are the cost and constraint matrices, respectively, and $N_A=N_e+N_l$. The matrix $\vc{A}_0$ enforces the homogenization variable through the constraint ${\vc{x}[1]}^2=1$. We call the constraints in~\eqref{eq:qcqp-matrix} the \textit{primary constraints}. 

\begin{example}[stereo-1D, known matrices] The cost and constraints matrices for the toy stereo problem are zero except for $\vc{Q}[2+i,2+i]=1$ and $\vc{Q}[1,2+i]=\vc{Q}[2+i,1]=-u_i$ for $i\in[N]$, $\vc{Q}[1,1] = \sum_i u_i^2$, $\vc{A}_i[1,2+i]=\vc{A}_i[2+i,1]=-m_i$, $\vc{A}_i[2,2+i]=\vc{A}_i[2+i,2]=1$, and $\bm{A}_i[1,1]=-2$, for $i>0$.
\end{example}

Problem~\eqref{eq:qcqp-matrix} is a~\ac{QCQP}. Its solution space, defined by a set of polynomial equality constraints, defines a real algebraic variety, which is a central object of the field of algebraic geometry. This is by itself an active area of research, with methods existing for finding, for example, the minimal set of constraints to uniquely define a variety~\cite{Groebner}. For the proposed paper, no knowledge of these advanced concepts is required as we take a numerical approach rather than an algebraic approach to describe the varieties. For the interested reader, we do include some references to the algebraic geometry perspective in footnotes. 

Because~\eqref{eq:qcqp-matrix} is, in general, NP-hard to solve, a common strategy is to relax the problem to a~\ac{SDP} by introducing $\vc{X}:=\vc{x}\vc{x}^\top$, which can be enforced using $\vc{X}\succeq0$ and $\rank{(\vc{X})}=1$, where the semidefinite constraint is convex while the rank constraint is not. We can drop the rank constraint and solve the following standard~\ac{SDP}:
\begin{equation}
  \min_{\vc{X}\inS{N}} \{ \inner{\vc{Q},\vc{X}} \barr \inner{\vc{A}_0, \vc{X}}=1, \inner{\vc{A}_i, \vc{X}}=0, i\inindex{N_A}\},
  \tag{P}
  \label{eq:sdp-primal}
\end{equation}
which is also called the primal or rank relaxation of~\eqref{eq:qcqp-matrix}.

\subsection{Global Optimality \pol{and Duality Theory}}

The~\ac{SDP} problem can be used in several ways to make claims about the global optimality of candidate solutions. Let us denote by $\oc{X}$ the solution of~\eqref{eq:sdp-primal} and its associated cost by $p^\star:=\inner{\vc{Q}, \oc{X}}$. If \oc{X} has rank one, then it can be factored as $\oc{X}=\oc{x}\oc{x}^\top$ and $\oc{x}$ is the optimal solution to~\eqref{eq:qcqp-matrix} with $q^\star:=f(\oc{x})=p^\star$. This leads us to the first form of tightness used in this paper.
\begin{definition}[Rank-tightness of the \ac{SDP} relaxation] We call the~\ac{SDP} relaxation~\eqref{eq:sdp-primal} \emph{rank tight} if its optimal solution $\oc{X}$ has rank one.
\end{definition}

\ac{SDP}s also enjoy a well-understood duality theory, which makes them great candidates for \textit{optimality certificates}. The Lagrangian dual problem of~\eqref{eq:sdp-primal} is given by
\begin{equation}
  d^\star = \max_{\rho, \vc{\lambda}} \{ -\rho \barr \vc{H}(\rho, \vc{\lambda}):=\vc{Q} + \rho\vc{A}_0 + \sum_{i=1}^{N_A}\lambda_i\vc{A}_i \succeq 0 \},
  \tag{D}
  \label{eq:sdp-dual}
\end{equation}
where $\rho$, $\vc{\lambda}={[\lambda_1, \cdots, \lambda_{N_A}]}^\top\inR{N_A}$ are the Lagrangian dual variables corresponding to $\vc{A}_0$ and $\vc{A}_i, i\inindex{N_A}$, respectively. It is well known that we always have $d^\star \leq p^\star \leq q^\star$. In what follows, we will also make the assumption that $d^\star=p^\star$, which holds under common constraint qualifications such as \textit{Slater's condition}~\cite{boyd_convex_2004} \pol{(see Figure~\ref{figure1})}.

We can use the dual problem to, instead of solving the primal \ac{SDP} and checking the rank of the solution, certify a local candidate solution \hc{x}. Indeed, using the \ac{KKT} conditions of~\eqref{eq:sdp-dual}, it is well-known (see~\eg,~\cite{cifuentes_local_2022}) that a solution candidate \hc{x} is globally optimal if there exist $\hat{\rho}, \hc{\lambda}$ such that
\begin{subequations}
  \begin{empheq}[left={\empheqlbrace\,}]{align}
    &\vc{H}(\pol{\hat{\rho}}, \hc{\lambda}) \hc{x} = \vc{0},\label{eq:stationarity} \\
    &\vc{H}(\pol{\hat{\rho}}, \hc{\lambda}) \succeq 0.\label{eq:dual-feas}
\end{empheq}\label{eq:certificate}
\end{subequations}
If these two conditions hold, we have \textit{strong duality}, meaning that $d^\star=p^\star=q^\star$ (right plot of Figure~\ref{figure1}). If we do not have strong duality, the above conditions cannot be jointly satisfied and we cannot \pol{use them to claim} global optimality of a candidate solution. Therefore, we introduce the notion of \textit{cost tightness}, a weaker form of tightness than rank tightness,\footnote{It is straightforward to see that rank tightness implies cost tightness.} which allows for candidate solutions to be certified:
\begin{definition}[Cost-tightness of the \ac{SDP} relaxation] We call the~\ac{SDP} relaxation~\eqref{eq:sdp-primal} \emph{cost tight} if $d^\star=p^\star=q^\star$. \end{definition}

Both forms of tightness may be useful in practice: when we have rank tightness, we can solve the~\ac{SDP} and derive the optimal value of the \ac{QCQP} from it. When the \ac{SDP} is prohibitively large, or when only cost tightness is attained, one may instead resort to a local solver and certify the solution candidate using Lagrangian duality. For completeness, we also mention that in some cases, one may extract a solution estimate from a higher-rank solution of the~\ac{SDP} in a procedure called ``rounding'', see~\eg,~\cite{rosen_se-sync_2019}. This typically consists of extracting the dominant eigenvector from $\vc{X}^\star$, and projecting it to the feasible set of~\eqref{eq:original}. Note that in this case there are \pol{not usually} guarantees on the quality of the solution and cases have been reported where the obtained estimate may be far from optimal~\cite{beck_exact_2008}.   

We have seen that either rank or cost tightness are necessary for efficiently obtaining or certifying globally optimal solutions, respectively. The remaining question is how one may increase the tightness of a given problem. This leads to the notion of redundant constraints, as explained next.

\subsection{Redundant Constraints}

Redundant constraints can be added to~\eqref{eq:qcqp-normal} \pol{and, equivalently,~\eqref{eq:qcqp-matrix}}, without changing its feasible set\pol{.}\footnote{Speaking in terms of algebraic geometry, the redundant constraints do not change the algebraic variety that is defined by the feasible set.} 
While the constraints are redundant for the~\ac{QCQP}, they may, however, change the feasible region of the~\ac{SDP}. In particular, redundant constraints typically reimpose structure on~\vc{X} that is lost when relaxing the rank-one constraint. For example, if the lifted vector is $\vc{x}^\top=\bmat{1 & \theta & \theta^2 & \theta^3}$, then 
\begin{equation}
  \vc{X} = \vc{x}\vc{x}^\top=\bmat{1 & \theta & \theta^2 & \theta^3 \\ \star & \theta^2 & \theta^3 & \theta^4 \\ \star & \star & \theta^4 & \theta^5 \\ \star & \star & \star & \theta^6},
  \label{eq:X}
\end{equation}
\noindent which has a very clear structure (it is a Hankel matrix~\cite{sdp_book}) that might be lost in the relaxation. The lifting constraints (in this case, $\vc{x}[3]={\vc{x}[2]}^2$ and $\vc{x}[4]=\vc{x}[3]\vc{x}[2]$) and symmetry of the solution take care of constraining all terms \pol{$\theta$, $\theta^2$, $\theta^3$, and} $\theta^5$, but nothing directly enforces that the elements corresponding to $\theta^4$ in the variable \vc{X} are equal. We can add the redundant constraint corresponding to (${\vc{x}[3]}^2=\vc{x}[2]\vc{x}[4]$) to enforce this. Redundant constraints can often be hard to find --- as the next example illustrates.

\begin{example}[stereo-1D, redundant constraints] A simple computation shows that 
  \begin{equation}
    z_i - z_j = \frac{1}{\theta - m_i} - \frac{1}{\theta-m_j} = (m_i - m_j)z_iz_j,
    \label{eq:ex-redundant}
  \end{equation}
  which holds for any \pol{$i,j$} and $z_i$, $z_j$ constructed using the lifting functions $\ell_i(\theta)$ introduced in~\eqref{eq:ex-lifting}. This shows that equation~\eqref{eq:ex-redundant}, which is quadratic in the elements of $\vc{x}$, is redundant in~\eqref{eq:qcqp-normal}, but non-redundant in the \ac{SDP}. It can be added with matrices $\vc{A}_{ij}$ with only non-zero entries $\vc{A}_{ij}[1,i+2]=\vc{A}_{ij}[i+2,1]=1$, $\vc{A}_{ij}[1,j+2]=\vc{A}_{ij}[j+2,1]=-1$, $\vc{A}_{ij}[i+2,j+2]=\vc{A}_{ij}[j+2,i+2]=(m_j-m_i)$, for all $i, j\inindex{N}, i\neq j$.
\end{example}

Because they impose more structure on \vc{X}, redundant constraints may have the effect of reducing the rank of \vc{X}, and thus improve the tightness of the relaxation. However, finding the right form and number of redundant constraints can be a tedious process, especially as the dimension of the problem increases. The present paper circumvents this process by proposing a numerical method to find all available redundant constraints, as we explain next.

\begin{figure*}[t]
  \centering
  \includegraphics[width=\linewidth]{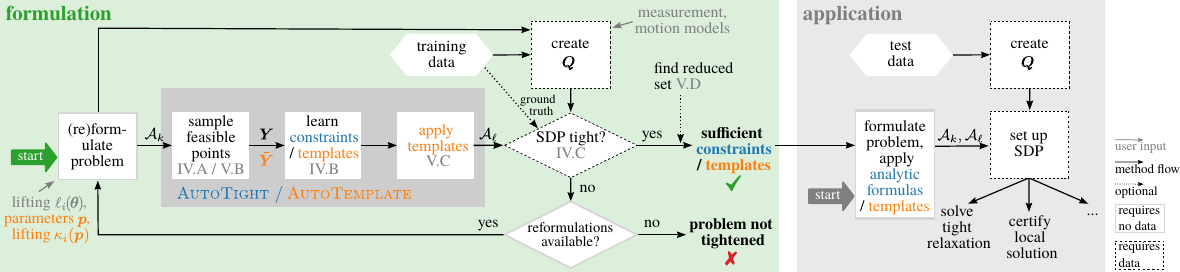}
  \caption{Overview of \autotight~and \autoscale. Using randomly generated training data, we find a set of sufficient redundant constraints or templates to tighten the relaxation, if it exists. We call this the \textit{formulation} phase, and it only has to be performed once for each problem type. The found templates (\autoscale) or interpreted constraints (\autotight) can be then used in the \textit{application} phase to solve or certify new test problems to global optimality, as long as the domain gap from training to test data is sufficiently small.}\label{fig:algo-patterns}
\end{figure*}

\section{\autotight}\label{sec:tightness}

In this section, we present \pol{our} method to determine whether a given \last{semidefinite relaxation can be tightened. We summarize the algorithm \autotight~in Figure~\ref{fig:algo-patterns}. In the \emph{formulation phase}, run only once for each new problem type using a representative example problem, all possible redundant constraints are automatically found and the tightness of the relaxation is evaluated. In the \emph{application phase}, the learned constraints are used to either solve or certify new problems created using incoming measurement data.}\footnote{\last{
During the formulation, ground truth information is usually available and can optionally be used to find the globally optimal solution efficiently. In the application phase, however, no ground truth information is required.}}

\subsection{Setting up the Nullspace Problem}\label{sec:tightness-nullspace}

At the core of the presented method is the insight that all of the \pol{possible} constraint matrices \pol{lie} in the nullspace of the linear subspace spanned by the feasible points of problem~\eqref{eq:qcqp-matrix}. We can thus determine constraints automatically by numerically retrieving the nullspace basis of a particular matrix. \pol{We will sometimes call these the \emph{learned constraints}.} The fact that feasible points do, indeed, form a linear space once lifted into a higher-dimensional space is proven in Appendix~\ref{app:thm1}.  

Assume we can generate feasible samples \pol{$\vc{\theta}^{(s)}$, $s\inindex{N_s}$, of~\eqref{eq:original}, and therefore also a set of lifted samples $\vc{x}^{(s)}$, constructed} using the \textit{known} substitutions $z_i^{(s)}=\ell_i(\vc{\theta}^{(s)})$:\footnote{Note that we can also allow for unknown  lifting functions, as long as a sampler of \vc{x} is available.} 
$\vc{x}^{(s)\top}=\bmat{\hom & \vc{\theta}^{(s)\top} & z_1^{(s)} & \cdots & z_{N_l}^{(s)}}.$
Then, for any valid constraint matrix $\vc{A}_i$ ($i>0$) of \eqref{eq:qcqp-matrix} (whether primary or redundant), we must have (by definition):
\begin{equation}
  \inner{\vc{A}_i,\vc{X}^{(s)}}= \vech{\vc{A}_i}^\top \vech{\vc{X}^{(s)}} = 0,
\end{equation} 
with $\vc{X}^{(s)}:=\vc{x}^{(s)}{\vc{x}^{(s)}}^\top$. This must hold for all samples $\vc{x}^{(s)}$. Defining the data matrix $\vc{Y}=\bmat{\vech{\vc{X}^{(1)}} & \cdots & \vech{\vc{X}^{(N_s)}}}\inR{n\times N_s}$, \pol{an} \pol{admissible} set of learned constraints, $\mathcal{A}_l$, is \pol{found in} the left nullspace basis of \vc{Y}:
\begin{equation}
  \mathcal{A}_l = \{ \vc{A}_1, \ldots, \vc{A}_{N_n} \} = \{ \vechinv{\vc{a}_i} \barr \vc{a}_i^\top\vc{Y}= \vc{0} \}.
  \label{eq:A-learned}
\end{equation}
In other words, each nullspace basis vector corresponds to one (vectorized) constraint matrix\pol{, and} finding all possible constraints is a standard nullspace problem. The dimension of the nullspace, $N_n$, corresponds to the total number of constraints. Note that we have exploited the fact that $\vc{X}^{(s)}$ and $\vc{A}_i$ are symmetric by using the half-vectorization operator, which reduces the problem size to $n:=\frac{N(N+1)}{2}$. 

By definition, all the constraints found through~\eqref{eq:A-learned} are linearly independent when operating in matrix form. When reformulating the constraints as a function of $\theta$ and adding them to~\eqref{eq:original}, however, some constraints may become dependent; in other words, the method finds both primary and redundant constraints. We explain this process in our pedagogical example at the end of this section, and in real-world applications in Section~\ref{sec:applications}.

Sometimes, it may be desirable to enforce some of the basis vectors to be known, for example to enforce that the original constraints from~\eqref{eq:qcqp-matrix} appear in the final formulation. 
We denote the set of \pol{known} constraints to be enforced by $\mathcal{A}_k=\{\widetilde{\vc{A}}_{1}, \ldots, \widetilde{\vc{A}}_{N_k}\}$. Completing the nullspace basis is as simple as appending the known constraints to the data matrix:
\begin{equation}
  \begin{aligned}
    \vc{Y} = \big[ \vech{\vc{X}^{(1)}} \, \cdots \, \vech{\vc{X}^{(N_s)}} \,\, \widetilde{\vc{a}}_{1} \, \cdots \, \widetilde{\vc{a}}_{N_k}\big],
\end{aligned}\label{eq:known}
\end{equation}
with $\widetilde{\vc{a}}_i=\vech{\widetilde{\vc{A}}_i}$. By definition, the left nullspace vectors of ${\vc{Y}}$ will then be orthogonal to the known constraints. 

To find a valid nullspace basis, we need to have at least $r=n-N_n$ samples, with $n$ the number of rows of $\vc{Y}$, $N_n$ the nullspace dimension, and $r$ the rank of $\vc{Y}$. However, since $r$ is not known a priori, a viable strategy is to randomly generate $N_s > n$ samples. This ensures that the data matrix is rank-deficient with probability one (because it has more columns than rows), The nullspace basis can then be reliably calculated using the permuted QR decomposition, as we explain next.\footnote{The procedure is equivalent to Algorithm 3 in~\cite{cifuentes_sampling_2017} and ensures \textit{poisedness}, as defined in Appendix~\ref{app:thm1}. Intuitively speaking, poisedness ensures that properties derived for the samples hold for the entire variety.}

\subsection{Sparse Basis Retrieval}\label{sec:tightness-qr}

Constraint matrices are expected to be sparse, since they usually involve only a subset of variables. \pol{Sparsity is good not only for lower runtime and memory consumption of \ac{SDP} solvers, but also for interpretability of the matrices. F}inding the sparsest nullspace basis is a NP-hard problem~\cite{coleman_null_1986}. However, when using a pivoted, or rank-revealing, QR decomposition~\cite{golub_matrix_2003} to find the left nullspace of the data matrix, we found the resulting constraints to be highly sparse. We show in Section~\ref{sec:results} that sometimes basis vectors are even as sparse as analytically identified constraints. Other matrix decomposition methods, such as the~\ac{SVD}, were empirically found to exhibit less sparsity.

The pivoted QR decomposition returns a decomposition of the form~\cite{golub_matrix_2003}
\begin{equation}
  \vc{Y}^\top \vc{P} = \pol{\vc{S}} \vc{R} = \pol{\vc{S}} \bmat{\vc{R}_1 & \vc{R}_2 \\ \vc{0} & \vc{0}},
  \label{<+label+>}
\end{equation}
where $\vc{P}$ is a $n\times n$ permutation matrix ensuring that the diagonal of $\vc{R}$ is non-increasing, \pol{\vc{S}} is $N_s \times N_s$ and orthogonal, $\vc{R}_1$ is upper-diagonal with dimensions $r \times r$, and $\vc{R}_2$ is of size $r \times N_n$. The nullspace basis vectors $\vc{a}_i$ are then given by
\begin{equation}
  \bmat{\vc{a}_1 & \cdots & \vc{a}_{N_n}} = \vc{P}\bmat{\vc{R}_1^{-1}\vc{R}_2 \\ -\vc{I}_{N_n}}.
  \label{eq:nullspace}
\end{equation}
Note that when using the permuted QR decomposition, the obtained basis vectors are linearly independent, but not necessarily orthogonal to each other, as would be the case with an~\ac{SVD}, for example. However, we found that the increased sparsity was of higher importance, both for computational speed and interpretability, than orthogonality.\footnote{By construction, each basis vector has at least $N_n-1$ zeros because of choosing the identity matrix in~\eqref{eq:nullspace}. \pol{The enforced sparsity in a few elements} was empirically found to also induce sparsity in the \pol{other} elements.} 

\subsection{Determining Tightness}\label{sec:tightness-sdp}

The method so far is independent of the cost function and only depends on the substitutions and primary constraints. To determine if the relaxation is tight, we consider a randomly generated problem setup, \pol{which defines} the cost matrix $\vc{Q}$ in~\eqref{eq:qcqp-matrix}. We determine tightness for this example setup \last{in the formulation phase}, as we explain next. Thanks to so-called~\ac{SDP} stability, we know that if the problem is tight, \pol{it will also be} tight for similar problems~\cite{cifuentes_local_2022} \last{in the application phase}, for example for similar noise and sparsity patterns.\footnote{Note that whenever we find a certifiably optimal solution for \pol{in the application phase}, we know that our test problem is `similar enough'. In other words, we have an a posteriori generalization guarantee.}

We determine cost tightness by comparing the optimal dual cost with the cost of a candidate global solution. The optimal dual cost is obtained by solving~\eqref{eq:sdp-dual} after adding all redundant constraints. The candidate global solution \last{can be} found by running an off-the-shelf local solver initialized at the ground-truth state, which we expect to be close to the optimal solution for low-enough noise. Indeed, this strategy allowed us to find the global minimum almost always for the noise regimes considered in Section~\ref{sec:results}.\footnote{If this fails \last{or no ground truth information is available, we can regenerate random setups and initializations} until we find that the cost of the candidate solution is equal to the dual cost (up to numerical tolerance). Then we know that it corresponds to the global minimum, because of duality theory~\cite{boyd_convex_2004}.} We compute the~\ac{RDG} between the cost of this local solution $\hat{q}$ and the optimal dual cost $d^\star$ through $(\hat{q}-d^\star)/\hat{q}$, and report cost tightness if the \ac{RDG} is below a fixed threshold (see Section~\ref{sec:hyper}). To determine rank tightness, we calculate the eigenvalues of the solution $\vc{X}$, and take the ratio of the first to the second-largest eigenvalue, called the~\ac{ER} in what follows. If the ratio is larger than a fixed value (see Section~\ref{sec:hyper}) we report that the solution is rank one. 

\subsection{Possible Outcomes}\label{sec:tightness-summary}

There are three possible outcomes of \autotight:
\begin{enumerate}
  \item The problem cannot be tightened.  Either a new formulation can be tried --- adding for instance (a subset of) higher-order monomial (Lasserre) terms~\cite{lasserre} --- or the non-tight \ac{SDP} can be used in conjunction with rounding, for example as an initialization for a local solver. 
  \item The problem is already tight or can be made so with few interpretable redundant constraints. We will see such a case in our pedagogical example and~\ac{RO} localization. In this case, constraints matrices can be efficiently created analytically and \autoscale is not necessary.
  \item The problem can be tightened, but only with non-interpretable redundant constraints. In this case, \autotight would have to be reapplied to every new problem instance, which does not scale to problem sizes typically encountered in robotics. We later present \autotemplate, which finds constraint \textit{templates} ---  patterns that can be applied to problems of any size.
\end{enumerate}
We conclude this section by studying the outcome of \autotight for our example problem.

\begin{example}[stereo-1D, \autotight] 
  We test tightness with $N=2$ landmarks. The randomly generated setup is $m_1=0.5488$ and $m_2=0.7152$, $u_1=18.14$, $u_2=-8.719$, and ground truth $\theta_{\text{gt}}=0.6028$.  \\
Since~\eqref{eq:ex-original1} is unconstrained, it is straightforward to construct feasible samples $\theta^{(s)}$ by uniform sampling, for example from the bounding interval of the anchor locations $m_i$. For each sample $\theta^{(s)}$, we create $\vc{x}^{(s)}$ using~\eqref{eq:ex-lifting}, and $\vech{\vc{X}^{(s)}}$ which is of size $n=10$. We choose $N_s\geq12$ samples, which is $10\%$ more than necessary, to create $\vc{Y}$. We perform a QR decomposition and observe that the nullspace is of dimension $N_n=3$. Converting the three basis vectors to matrices, we get as elements of $\mathcal{A}_\ell$: 
  \begin{equation}\label{eq:ex-a_list}
  \footnotesize{
  \bmat{0 & 0  & -\alpha & 0 \\
        \star  & 0  & -1    & 1  \\
        \star  & \star   & 0     & -2\gamma \\
        \star  & \star   & \star      & 0
        },
  \bmat{0 & 0  & 1 & -1 \\
        \star  & 0  & 0    & 0  \\
        \star  & \star   & 0  & \alpha \\
        \star  & \star   & \star      & 0
        },
        \bmat{1 & 0  & \beta & 0 \\
        \star  & 0  & 0    & -0.5  \\
        \star  & \star   & 0  & \gamma \\
        \star  & \star   & \star      & 0
        }
   }
\end{equation}%
\noindent where $\alpha$, $\beta$ and $\gamma$ are numeric values -- we find by inspection that $\alpha=m_2-m_1$, $\beta=\frac{m_2}{2}$, and $2\gamma=m_2(m_2-m_1)$. \\
We evaluate the tightness by solving~\eqref{eq:ex-original1} using~\ac{GN} initialized at ground truth on one hand (yielding $\hat{\theta}$ and $\hat{q}$), and the~\ac{SDP} relaxation~\eqref{eq:sdp-primal} on the other hand (yielding $\vc{X}^\star$ and ${p}^\star$). We find $\hat{\theta}=0.6038, \hat{q}=0.06857$, and an~\ac{RDG} of $1.6\times10^{-6}$. The rank of $\vc{X}^\star$ is two, and the problem is thus cost tight but not rank tight. 
However, when imposing only the known substitution constraints and no redundant constraints, we have an~\ac{RDG} of $1.0$ and thus neither cost nor rank tightness. \\
We conclude that this problem formulation has a tight relaxation with redundant constraints. For this example, in order to scale to larger problems, we can interpret the constraints~\eqref{eq:ex-a_list} or use directly the analytic form~\eqref{eq:ex-redundant} (we are thus in the second outcome in Section~\ref{sec:tightness-summary}). Note, however, that even for this simple problem, interpreting~\eqref{eq:ex-a_list} was not trivial. For more complex problems, we thus propose to use \autotemplate, which we introduce in the next section.
\end{example}

\subsection{Theoretical Guarantee}\label{sec:tightness-thm}

The success of the \autotight algorithm provides a trivial sufficient condition that a given problem formulation can be tightened with redundant constraints. In this section, we show the converse of this statement: the formulation can be tightened \textit{only if} \autotight is successful. That is, the \autotight method is \emph{complete}, \pol{as formalized in Theorem~\ref{thm:1}.}

\begin{theorem}[Completeness of \autotight]\label{thm:1}
  Let $\mathcal{A}_{\ell}$ be the set of learned redundant constraints using \autotight, and let $\mathcal{A}_{k}$ be any other set of redundant constraints (perhaps found via a different method). Let (P$_{\ell}$) and (P$_k$) be the SDP~\eqref{eq:sdp-primal} constructed from $\mathcal{A}_{\ell}$ or $\mathcal{A}_k$, respectively. Denote by $(\vc{X}_{\ell}^\star,p_{\ell}^\star)$ and $(\vc{X}_k^\star,p_k^\star)$ the optimal values and costs of (P$_\ell$) and (P$_k$), respectively, and by $q^\star$ the optimal cost of~\eqref{eq:original}. Then we have: \begin{equation}
    \begin{aligned}
      q^\star=p_k^\star &\implies q^\star=p_\ell^\star,  \\
      \rank{(\vc{X}_{k}^\star)}=1 &\implies \rank{(\vc{X}_{\ell}^\star)}=1. \\
    \end{aligned}
    \label{eq:thm1}
  \end{equation}
\end{theorem}

The proof of Theorem~\ref{thm:1} uses results from Cifuentes~\etal~\cite{cifuentes_sampling_2017} and is provided in Appendix~\ref{app:thm1}. The implication of Theorem~\ref{thm:1} is that if \autotight does not lead to a tight relaxation of a given problem formulation, then there is no other set of constraints that can tighten the problem, and one needs to instead reformulate the problem, \eg, ascend in Lasserre's hierarchy~\cite{lasserre}.

\section{\autotemplate}\label{sec:scalability}

The method \autotight determines whether a given problem can be tightened, but unless the learned constraints are interpretable, we need to repeatably apply \autotight as we change the problem formulation. This induces unnecessary computational overhead and may be prohibitively expensive for large problem sizes. We thus present \autoscale, a scalable extension of \autotight.

\subsection{Setting up the Nullspace Problems}\label{sec:nullspace}

While problems in robotics tend to be high-dimensional, they usually only exhibit a few different variable types. For example, in~\ac{SLAM}, the only variable types are robot poses and landmark positions~\cite{barfoot_state_2017}. We would expect the constraints relating instances from the same variable types to be repeatable; for example, all constraints that involve \pol{a single pose} should hold for all other poses too. \autoscale uses this insight to learn \emph{templates} that act on variable sets and can thus be more easily scaled.

To give a  simple example, when learning the constraints for the \ac{RO} localization problem (see for example the first row of Figure~\ref{fig:ro-matrices} in Section~\ref{sec:results-range}) using \autotight, the same constraint ($z_i=\norm{\vc{\theta}_i}^2$) is found three times: once for each position. Using \autoscale, we learn only one \emph{template} of this form, and can then apply it to as many positions as necessary in the application phase.

Given a problem formulation, we first identify the different variable types \pol{and a sequence of variable sets}. Variable types can be landmarks, poses, substitutions, \etc.  
\pol{Each variable set represents one group of variables for which we learn constraints.} A simple recipe for creating variable sets goes as follows. We start by creating one variable set per variable type, including one instance and the homogenization variable per set. Then, we incrementally add one more variable at a time, using always all possible permutations of variable types. The resulting variable sets, for the applications considered in this paper, are given in Table~\ref{tab:tight-problems} in Section~\ref{sec:results}. We call each variable set $\mathcal{G}_k$ and write ${(\cdot)}_{\mathcal{G}_k}$ to reduce a variable or function to the elements included in $\mathcal{G}_k$.

\begin{example}[stereo-1D, variable sets]
For our pedagogical example, the variable types are the position $\theta$, the substitutions $z_i$, and $h$. Note that the landmarks are known and not considered as variables. The above recipe then leads to the following \pol{list} of variable sets: $[\{\hom, \theta\}\}, \{\hom, z_1\}, \{\hom, \theta, z_1\},\,$ $\{\hom, z_1, z_2\}, \{\hom, \theta, z_1, z_2\}, \{\hom, z_1, z_2, z_3\}, \{\hom, \theta, z_1, z_2, z_3\},\cdots ]$.
\end{example}

\subsection{Factoring Out Parameters}\label{sec:parameters}

For some problems, constraints may depend on parameters that are known a priori. In the stereo-1D example and in the stereo localization problem presented in Section~\ref{sec:setup-stereo}, for example, the constraints depend on the known landmark coordinates. Because of this, the learned templates may be harder to interpret, and not applicable to other random setups. To overcome this problem, we treat such quantities as \textit{parameters} and append them to the samples of feasible points. Let $\vc{p}\inR{N_p}$ be the vector of $N_p$ parameters\last{, chosen rich enough so that all constraints are linear functions of $\vech{\vc{X}^{(s)}} \otimes \vc{p}$. We always set $\vc{p}[1]=1$} to include the case without added parameter dependencies.\footnote{Note that defining the parameters \last{requires user input: for example, for a problem with landmarks, the user needs to decide if \vc{p} contains only the landmark coordinates or also high-order monomials thereof (see the stereo-1D example or Section~\ref{sec:stereo-form} for more concrete examples). This is still significantly simpler than manually finding all required redundant constraints, and comparable to the choice of a sparse Lasserre's hierarchy, for which systematic methods exist that could likely be employed here as well~\cite{sparse-Lasserre}.}}

For scalability, we do not add all elements of \last{$\vc{p}$} for each considered variable set, but only the ones that depend on the variables in $\mathcal{G}_k$, which we denote by $\vc{p}_{\mathcal{G}_k}$. For a visualization of these concepts, we return to the pedagogical example.

\begin{example}[stereo-1D, augmented variable sets] This  problem has parameters $m_i$. Because the lifting constraints $l_i(\vc{x})$ are linear in $\theta$ and $m_i$, respectively, we use \last{$\vc{p}=\bmat{1 & m_1 & \cdots & m_N}^\top$}. For this particular example, we know the redundant constraints are of the form~\eqref{eq:ex-redundant}, which confirms that this choice of parameters is rich enough: each redundant constraint can be written as a linear combination of the elements of $\vech{\vc{X}^{(s)}} \otimes \last{\vc{p}}$.
\end{example}

We modify each feasible sample to include its parameter dependencies, leading to the \emph{augmented} feasible sample
\begin{equation}
  \bar{\vc{z}}^{(s)}_{\mathcal{G}_k} := \vech{\vc{X}_{\mathcal{G}_k}^{(s)}} \kron \last{\vc{p}_{\mathcal{G}_k}} \inR{\bar{n}_k},
  \label{eq:x-augmented}
\end{equation}
with $\bar{n}_k:=n_k K_k$, $n_k$ the dimension of $\vech{\vc{X}_{\mathcal{G}_k}}$, and $K_k$ is the dimension of \last{$\vc{p}_{\mathcal{G}_k}$}. We get the augmented data matrix  
\begin{equation}
  \bar{\vc{Y}}_{\mathcal{G}_k} = \bmat{\bar{\vc{z}}_{\mathcal{G}_k}^{(1)} & \cdots & \bar{\vc{z}}_{\mathcal{G}_k}^{(\bar{N}_s)}}\inR{\bar{n}_k \times \bar{N}_s}, 
  \label{eq:Y-augmented}
\end{equation}
where the number of samples $\bar{N}_s$ has to be chosen to ensure that $\bar{\vc{Y}}_{\mathcal{G}_k}$ is rank-deficient{, as in Section~\ref{sec:tightness-nullspace}}. We denote the left nullspace basis vectors of~\eqref{eq:Y-augmented} by $\bar{\vc{a}}_l\inR{n_k K_k}$, with $l\inindex{\bar{N}_n}$. We call these basis vectors \textit{templates} \pol{and we will apply them to new problems as explained next.}

\subsection{Applying Templates}\label{sec:applying}

Conceptually speaking, applying the templates means repeating each constraint for each possible combination of the variables that it involves. For example, if one constraint matrix involves one position and two different landmarks, then we repeat the constraint for each position and each possible pair of landmarks per position. To facilitate this operation programmatically, we have created \pol{a tool} to generate sparse matrices using variable names for indexing.\footnote{The code is available as a \pol{stand-alone} open-source package at~\url{https://github.com/utiasASRL/poly_matrix}.} That way, applying constraints to all possible variables \pol{of a given type} simply means creating duplicates of a given constraint, and then renaming the variables that it involves. 

If parameters were factored out as explained in Section~\ref{sec:parameters}, then they need to be factored back in before solving the~\ac{SDP}, using the current parameter realization. We introduce the operator $\mataug{\cdot}$, which folds the augmented basis vector $\bar{\vc{a}}_l$  (which we recall has $n_kK_k$ dimensions, with $K_k$ the number of lifted parameters) column-wise into a $n_k\times K_k$ matrix. Then, \textit{factoring in} a parameter realization $\vc{p}^{(s)}$ can be written as

\begin{equation}
  \vc{a}_l = \mataug{\bar{\vc{a}}_l} \vc{p}^{(s)},
  \label{eq:a-small}
\end{equation}
where the output $\vc{a}_l\inR{n_k}$ is now a problem-specific vectorized constraint that can be converted to the corresponding constraint matrix $\vc{A}_l=\vechinv{\vc{a}_l}$. \pol{Zero padding is used to apply constraints learned on subsets of variables for the full variable set}.
We return to the stereo-1D example to illustrate these concepts:

\newcommand{\Gf}{{\mathcal{G}_5}}
\begin{example}[stereo-1D, templates]
  We apply \autoscale to the stereo-1D example, imposing the known substitution constraints.  We do not find any additional constraints for all first variable sets. Only when using the group $\Gf=\{h, \theta, z_1, m_1, z_2, m_2\}$, each augmented sample is of the form:
  \begin{equation}
    \begin{aligned}
      \bar{\vc{z}}_\Gf^{(s)\top} =\,\,\, 
      \big[ & 1 \,\,\, \theta \,\,\, z_1 \,\,\, z_2 \,\,\, \theta^2 \,\,\, \theta z_1 \,\,\, \theta z_2 \\
    & z_1^2 \,\,\, z_1z_2 \,\,\, z_2^2\big]^{(s)} \kron \bmat{1 & m_1 &m_2}^{(s)},
    \end{aligned}
  \end{equation}
The redundant constraints from~\eqref{eq:ex-redundant} are in the nullspace of the augmented data matrix and can thus be learned automatically. Indeed, the template corresponding to $\vc{A}_{12}$ can be written as
  \begin{equation}
    \begin{aligned}
    \bar{\vc{a}}_{12}^\top &= \bmat{{\vc{\alpha}}_1^\top & {\vc{\alpha}}_2^\top & {\vc{\alpha}}_3^\top}, \\
    &{\vc{\alpha}}_1^\top = \bmat{0 & 0 & 1 &\-1 & 0 & 0 & 0 & 0 & \,\,0 & 0}, \\
     &{\vc{\alpha}}_2^\top = \bmat{0 & 0 & 0 & \,\,0 & 0 & 0 & 0 & 0 & \,\,1 & 0}, \\
     &{\vc{\alpha}}_3^\top = \bmat{0 & 0 & 0 & \,\,0 & 0 & 0 & 0 & 0 &\-1 & 0}.
  \end{aligned}
  \end{equation}
  and satisfies $\bar{\vc{a}}_{12}^\top\bar{\vc{z}}_{\mathcal{G}_5}^{(s)}=0$ for any sample $\bar{\vc{z}}_{\mathcal{G}_5}^{(s)}$.
  Note that the template $\bar{\vc{a}}_{12}$ does not depend on the landmarks anymore. 
  Given new realizations of parameters $\vc{p}^{(s)}$, we can create $\vc{p}^{(s)}_{\mathcal{G}_5} = \big[ 1 \, m_1^{(s)} \,  m_2^{(s)}\big]^\top$ and the corresponding constraint matrix $\vc{A}_{12}^{(s)} = \vechinv{\vc{a}_{12}^{(s)}}$, with
  \begin{equation}\label{eq:ex-mataug}
    \vc{a}_{12}^{(s)} = \mataug{\bar{\vc{a}}_{12}} \vc{p}^{(s)}_{\mathcal{G}_5}
    = \big[\bm{\alpha}_1 \,\, \bm{\alpha}_2 \,\, \bm{\alpha}_3\big] \vc{p}^{(s)}_{\mathcal{G}_5}.
  \end{equation}
\end{example}

\subsection{Reducing the Number of Constraints}\label{sec:sparsity}

\pol{In practice,} not all of the found templates are actually  necessary for tightness. Therefore, we suggest to prune the found templates before applying them to large problem sizes. Assume we have found a set of learned constraints $\mathcal{A}_l$ for which the problem is (at least) cost tight. Then, we can solve the following optimization problem in an attempt to sort the constraints by their importance for tightness: 
\begin{equation}
  \begin{aligned}
    \min_{\vc{\lambda}, \rho}&  \norm{\vc{\lambda}}_1 \\
    \text{s.t.} \quad & \vc{H}(\rho, \vc{\lambda}) \succeq 0  \\
    & \vc{H}(\rho, \vc{\lambda})\hc{x} = \vc{0}, \\
  \end{aligned}
  \label{eq:sparsity}
\end{equation}
where $\norm{\cdot}_1$ denotes the $L_1$-norm, $\vc{H}$ is defined as in~\eqref{eq:sdp-dual} (with the learned matrices substituted for $\vc{A}_k$)  and $\hc{x}$ is the optimal solution of~\eqref{eq:qcqp-matrix}. Intuitively, Problem~\eqref{eq:sparsity} finds a sparse set of dual variables required for cost tightness, as the $\ell^1$-norm promotes sparsity. By ordering the learned constraints by decreasing magnitude of $\vc{\lambda}$ and adding them one by one, we find a smaller subset of constraints that is also sufficient for cost tightness. This problem naturally lends itself to a bisection-like algorithm, where we try using all and no redundant constraints at first, and then continue to try cutting the interval in half. We terminate when the considered interval is of size one. \pol{Using} only these constraints as templates significantly reduces the computational cost of all downstream operations, as shown in Section~\ref{sec:results}. We call this smaller set of constraints, empirically found to be also sufficient for tightness, the \textit{reduced constraints}.

As another pruning step, we also make sure that all constraints are linearly independent after applying templates to other variables. For this purpose, we use the same rank-revealing QR decomposition as in Section~\ref{sec:tightness-qr} but keep only the valid range-space basis vectors. Because of the sparsity of the constraints, this adds no significant cost.   

\subsection{Summary}\label{sec:summary}

To summarize, \autoscale consists of the following steps, as displayed in Figure~\ref{fig:algo-patterns}. \last{During the formulation phase}, we set up the nullspace problem for one variable set at a time, including corresponding parameters, as explained in Sections~\ref{sec:nullspace} and~\ref{sec:parameters}. This allows us to learn templates instead of constraints. \pol{Before checking tightness with the same procedure as for \autotight (Section ~\ref{sec:tightness-sdp})}, we apply these templates to all other variables of the same \pol{variable} \pol{group} as outlined in Section~\ref{sec:applying}. If the problem is tight, we return the learned templates, after optionally extracting a smaller sufficient subset as explained in Section~\ref{sec:sparsity}. If the relaxation is not tight, we either restart the process, using the next variable set if available, or go up in Lasserre's hierarchy if all groups at this level have been exhausted. \last{During the application phase, we apply the templates to new problem setups of any size.}

In terms of complexity, one important computational load of both methods, \autotight and \autoscale, is the nullspace calculation, which involves a permuted QR decomposition and is cubic in complexity. The crucial dimensions are thus $n$ and $\max_k \bar{n}_k$, respectively, which correspond to the dimensions of the feasible samples in the respective methods. For \autotight, if higher-order substitutions are required for tightness, this $n$ grows quickly (see, for example, the results in Figure~\ref{fig:results-time-stereo}). For \autoscale, the dimensions $\bar{n}_k$ depend on the chosen variable sets and can be kept as small as possible by growing the size of the groups incrementally (using  for example our proposed recipe), until tightness is achieved. This is the main reason for better scalability of \autoscale. The application of the learned templates to new problem setups in \autoscale~may add to the computational load when the number of learned templates is large, as we will see in the stereo-localization application. 

\begin{table}[t]
  \caption{Overview of the tightened problems, including the variable sets, problem dimensions, and noise parameters. For simplicity, all substitutions are called $\vc{z}_i$. $N_{\rm out}$ \pol{is} the number of outliers, and noise \pol{corresponds} to the standard deviation of zero-mean Gaussian noise.}\label{tab:tight-problems}
  \footnotesize{
    \begin{tabularx}{\linewidth}{ll}
      \mc2{l}{\pol{Problem (Parameters; Inlier noise / Outlier noise): \textit{Variables}}}\\ 
\midrule
\mc{2}{l}{\ac{RO} loc. ($d=3, N_m=10, N=3$; $10^{-2}$):} \\ 
\mc{2}{R{.98\linewidth}}{$\pol{[}\{\hom, \vc{\theta}_2\}, \{\hom, \vc{z}_1\}, \{\hom, \vc{\theta}_1, \vc{z}_1\}, \cdots\pol{]}$} \\ 
\midrule
\mc2{l}{stereo loc. ($d\in\{2,3\},N=d+1$; $10^{0}$):} \\ 
\mc2{r}{$\pol{[}\{\hom, \vc{\theta}\}$, $\{\hom, \vc{z}_1\}, \{\hom, \vc{\theta}, \vc{z}_1\} , \{\hom, \vc{z}_1, \vc{z}_2\}, \cdots\pol{]}$} \\  
\midrule
\mc2{l}{PPR~\cite{briales_convex_2017} ($d=3, N=3$; $10^{-2}$), PLR~\cite{briales_convex_2017} ($d=3, N=5$; $10^{-3}$):} \\ 
\mc2{r}{$\pol{[}\{\hom, \vc{\theta}\}\pol{]}$} \\
\midrule
\mc2{l}{rPPR~\cite{yang_certifiably_2022} ($d=3, N=4, N_{\rm out}=1$;  $10^{-2} / 10^{0}$),}\\
\mc2{l}{rPLR~\cite{yang_certifiably_2022} ($d=3, N=6, N_{\rm out}=1$; $10^{-3} / 10^{-1}$):}\\ 
\mc{2}{r}{$\pol{[}\{\hom, \vc{\theta}\}, \{\hom, \vc{\theta}, w_1\}, \{\hom, \vc{\theta}, \vc{z}_1\}, \{\hom, \vc{\theta}, w_1,w_2\},$} \\
\mc{2}{r}{$\{\hom, \vc{\theta}, w_1, \vc{z}_1\}, \{\hom, \vc{\theta}, \vc{z}_1, \vc{z}_2\},\cdots\pol{]}$}
    \end{tabularx}
  }
\end{table}

\section{Applications}\label{sec:applications}

We study two real-world applications of the proposed method in depth: \ac{RO} and stereo localization. \pol{Their sketches and the corresponding} factor graph are provided in Figure~\ref{fig:setups}. We outline the problem formulations in this section, and report simulated and experimental results in the \pol{next} two sections. \pol{Formulations from the literature that we also study are provided in Appendix~\ref{app:math}.}

\subsection{Range-Only Localization}\label{sec:setup-range}

The first application is \ac{RO} localization with fixed and known landmarks, as encountered in~\ac{UWB}-based localization~\cite{mueller_fusing_2015,goudar_gaussian_2022-1} or 
WiFi- or Bluetooth-based indoor localization~\cite{zafari_survey_2019}. This problem has been \pol{widely} studied, \pol{for example,} in~\cite{beck_exact_2008,larsson_optimal_2019} and, more recently, in \cite{pacholska_relax_2020} and \cite{dumbgen_safe_2023}. We reproduce the tightness results from~\cite{dumbgen_safe_2023} using our new method, and then study a different formulation, which we automatically tighten using \autotight~and \autoscale. 

\subsubsection{Problem Statement}

\begin{figure}[t]
  \centering
  \includegraphics[width=\linewidth]{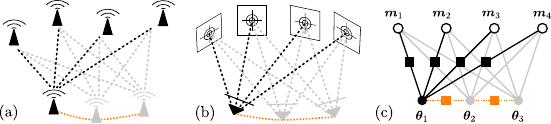}
  \caption{Sketches of the two example applications and their corresponding factor graphs: (a) \ac{RO} localization, (b) stereo camera localization and (c) the factor graph for both problems, where the filled circles are unknown states, the squares are landmark measurement factors, and the empty circles are known landmarks. For completeness, we plot in orange the motion-prior factors, which could be added as in~\cite{dumbgen_safe_2023} but are omitted here for simplicity.}
  \label{fig:setups}
\end{figure}

The goal of~\ac{RO} localization is to estimate the position of a moving device over time, given range measurements to fixed and known landmarks, also called anchors in this context. We call the anchors $\vc{m}_k\inR{d}$ with $k\inindex{N_m}$ and the position at time $t_n$ is denoted $\vc{\theta}_n\inR{d}$, with $n\inindex{N}$. We use $d=3$ in all of the experiments, and\label{short45} the following common formulation of the problem~\cite{beck_exact_2008}:
\begin{equation}
  \min_{\bm{\theta}} \,c(\bm{\theta}), \quad c(\bm{\theta})= \sum_{n, k \in\mathcal{E}} \left(d_{nk}^2 - \norm{\vc{m}_k - \vc{\theta}_n}^2\right)^2,
  \label{eq:ro-original}
\end{equation}
where $\mathcal{E}$ is the edge set of the measurement graph, with an edge between position $n$ and anchor $k$ if their distance $d_{nk}$ is measured, and $\vc{\theta}$ is the concatenation of unknowns~$\{\vc{\theta}_n\}_{n=1}^N$.\footnote{Note that it is straightforward to include a motion prior in~\eqref{eq:ro-original}, such as a constant-velocity prior, as shown in~\cite{dumbgen_safe_2023} and visualized in Figure~\ref{fig:setups}. Such priors are typically up to quadratic in the unknowns, thus not requiring any special treatment when it comes to constraints, and are omitted for simplicity.}

\subsubsection{Problem Reformulation}

Problem~\eqref{eq:ro-original} is quartic in the unknowns, and thus may contain multiple local minima~\cite{dumbgen_safe_2023}. However, by introducing lifting functions that are quadratic in $\vc{\theta}_n$, it can be written as a~\ac{QCQP}, making it a candidate for~\ac{SDP} relaxation. We study two such lifting functions. First, it is instructive to expand all elements of the cost:
\begin{equation}
  c(\bm{\theta}) = \sum_{n, k \in\mathcal{E}} \left(d_{nk}^2 - \norm{\vc{m}_k}^2 + 2 \vc{m}_k^\top \vc{\theta}_n - \norm{\vc{\theta}_n}^2\right)^2.
  \label{eq:ro-cost-out}
\end{equation}
Looking at~\eqref{eq:ro-cost-out}, we see that the substitution 
\begin{equation}
  z_n:= \norm{\vc{\theta}_n}^2\inR{}
  \label{eq:ro-sub1}
\end{equation}
is enough to make the problem quadratic in the lifted vector $\vc{x}^\top=\bmat{\hom &\vc{\theta}^\top& z_1 & \cdots & z_N}$. The same substitution was used in~\cite{dumbgen_safe_2023} and was shown to require no redundant constraints for tightness. This substitution~\eqref{eq:ro-sub2} is also called a \textit{sparse} Lasserre substitution~\cite{sparse-Lasserre} .  We also study the more methodological (dense) substitution that introduces all quadratic terms of the elements of $\vc{\theta}_n$:
\begin{equation}
  \vc{y}_n:= \vecaug{\vc{\theta}_n\vc{\theta}_n^\top}\inR{d(d+1)/2}.
  \label{eq:ro-sub2}
\end{equation}

In general, given polynomial cost and constraints functions of degree up to $k$, it suffices to introduce lifting functions up to degree $\left\lceil{\frac{k}{2}}\right\rceil$. Out of all possible lifting functions, some may not require redundant constraints (for example $z_n$ above), while others do (for example $\vc{y}_n$ above), which we will study in more detail in Section~\ref{sec:results-range}.

As in the pedagogical example, sampling feasible points is straightforward as the problem is initially unconstrained. We sample points uniformly at random from the convex hull of the anchors, yielding $\vc{\theta}^{(s)}$, and create $\vc{x}^{(s)}$ and $\vc{Y}$ as explained in~Section~\ref{sec:tightness-nullspace}.

\subsection{Stereo Localization}\label{sec:setup-stereo}

The second application is the estimation of the pose of a stereo camera by minimizing the reprojection error of known landmarks, which we refer to as stereo localization. The reprojection error can be used to model Gaussian noise on pixel measurements~\cite{matthies_error_1987}. To the best of our knowledge, this problem has not been successfully relaxed to a tight \ac{SDP} before, with common solutions typically resorting to the back-projection error~\cite{terzakis_consistently_2020,sun_certifiably_2020} (\ie, the error is assumed Gaussian in Euclidean space). Closest to our solution is~\cite{olsson_branch-and-bound_2009}, where a branch-and-bound method in combination with a (non-tight) semidefinite relaxation is used to minimize the reprojection cost. Instead, we use the proposed methods to 1) find a new formulation of the problem that can be tightened \pol{using} \autotight, and \pol{to} 2) generate templates that can be scaled to new problem instances \pol{using} \autoscale.

\subsubsection{Problem Statement}

Our goal is to estimate the pose of a stereo camera given the measured image coordinates, in both left and right frames, of a number of known landmarks. We call the known landmarks $\vc{m}_k$ and the homogenized form $\bar{\vc{m}}_k=\bmat{\vc{m}_k^\top&1}^\top$, with $k\inindex{N}$.\footnote{We use $N$ and not $N_m$ because in stereo localization, the number of landmarks determines the problem size $N$ (since the number of poses is fixed to one).} For simplicity, we focus on one measurement time only, and call the unknown pose at that time $\vc{T}\in\SE{d}$, which contains both the rotation matrix from world to camera frame, $\vc{C}\in \SO{d}$, and the associated translation $\vc{t}\inR{d}$. We collect the pixel measurements of landmark $k$ in $\vc{y}_k^\top := \bmat{u_k^{\ell} & v_k^{\ell} & u_k^r & v_k^r}$, where $u$ and $v$ denote the $x$ and $y$ coordinates in pixel space, and superscripts $\ell$ and $r$ correspond to the left and right frame, respectively. We call the intrinsic stereo camera matrix in $d$ dimensions $\vc{M}_d$, with for example:
\def\subk{\left(\vc{e}_d^\top\vc{T}\bar{\vc{m}}_k\right)^{-1}\vc{T}\bar{\vc{m}}_k}
\vspace{-1em}
\begin{equation}
\vc{M}_2=\bmat{f_u&c_u& \fub \\
f_u&c_u& \shortminus\fub},\,
  \vc{M}_3=\bmat{f_u&0&c_u&\fub\\0&f_v&c_v&0\\
  f_u&0&c_u& \shortminus\fub \\0&f_v&c_v&0},
  \label{eq:stereo-M}
\end{equation}
\noindent
where $b_u=f_u\frac{b}{2}$, and $f_u$, $f_v$, and $b$ are the focal lengths and baseline, respectively. 
Given pixel measurements from $N$ landmarks, the pose can be estimated as the solution of the optimization problem
\begin{equation}
  \min_{\vc{T}\in\SE{d}} c(\vc{T}), \,c(\vc{T}) =\sum_{k\inindex{N}} \|\vc{y}_k - \vc{M}_d\subk\|^2,
  \label{eq:stereo-original}
\end{equation}
where $\vc{e}_d$ is the $d$-th standard basis vector.

\subsubsection{Problem Reformulation}\label{sec:stereo-form}

Due to the \SE{d} constraint and the rational cost function, Problem~\eqref{eq:stereo-original} is hard to solve globally. However, the problem can again be lifted to a~\ac{QCQP} by introducing a series of relaxations and substitutions. First, we relax the $\SO{d}$ to a $\Orth{d}$ constraint, which essentially drops the $\det{\vc{C}}=1$ constraint. As discussed in~\cite{rosen_se-sync_2019}, this relaxation is often tight without additional constraints, and if not, handedness constraints can be added~\cite{briales_convex_2017}. As we are automatically finding all redundant constraints, these constraints will be added later if required. Secondly, by inspecting the cost in~\eqref{eq:stereo-original}, we note that the following substitution makes the cost quadratic: 
\begin{equation}
  \vc{v}_k= \subk, 
\end{equation}
and it can be enforced as quadratic constraints $\bm{l}_k(\bm{T})=\bm{0}$ by multiplying both sides by the denominator. We obtain the following~\ac{QCQP}:
\begin{equation}
  \begin{aligned}
    \min_{\vc{C},\vc{t}} &\sum_{k\inindex{N}} \|\vc{y}_k - \vc{M}_d\vc{v}_k\|^2 \\
    \text{s.t. } & \left(\vc{I}_d - \vc{v}_k\vc{e}_d^\top\right)\vc{T}\bar{\vc{m}}_k=\vc{0} \text{, } k\inindex{N} \\
    & \vc{C}^\top\vc{C} = \vc{I}_d.
\end{aligned}
  \label{eq:stereo-qcqp}
\end{equation}

\pol{Because} the $d$-th element of $\vc{v}_k$ is always one by definition, we introduce
\begin{equation}
  \vc{u}_k^\top := \bmat{\vc{v}_k[1] & \cdots & \vc{v}_k[d-1] & \vc{v}_k[d+1]}, 
  \label{eq:u}
\end{equation}
  and we write~\eqref{eq:stereo-qcqp} as a~\ac{QCQP} in the lifted vector
\begin{equation}
  \vc{x}^\top = \bmat{\hom &\vc{t}^\top & \ve{\left(\vc{C}\right)}^\top& \vc{u}_1^\top& \cdots& \vc{u}_{N}^\top}.
  \label{eq:stereo-x}
\end{equation}

For sampling feasible points, we randomly generate positions $\vc{t}^{(s)}$ uniformly from a unit cube, and orientations $\vc{C}^{(s)}$ from uniformly sampled quaternions. We construct $\vc{Y}$ from these samples as outlined in Section~\ref{sec:tightness-nullspace}.

When  using \autoscale, we first add parameters that are linear polynomials of each landmark's coordinates, because the substitutions result in constraints that are linear in both the landmark coordinates and the camera pose. However, we will see in Section~\ref{sec:results-stereo} that we need to add additional quadratic substitutions to achieve tightness. We therefore include all up to quadratic monomials in the \pol{parameters $\vc{p}$}.

\section{Simulation Results}\label{sec:results}

\pol{We are now equipped to apply \autotight and \autotemplate to the two example problems and problems from the literature. In this section, we demonstrate the performance on simulated data, while in the next section, we provide results on real-world datasets for \ac{RO} and stereo localization} 

\definecolor{LightGray}{RGB}{150,150,150}
\definecolor{LightRed}{RGB}{255,114,118}
\begin{table}[t]
  \caption{Overview of the considered problems, their tightness and whether there are redundant constraints. Highlighted in red are formulations that were found to be non-tight.}\label{tab:all-problems}
  \centering
  \begin{tabular}{lc|ccc}
    Problem & lifting & redundant constr. & cost tight & rank tight \\
    \midrule
    \rowcolor{LightGray!50}
    \ac{RO} loc.& z~\eqref{eq:ro-sub1} & no  & yes & yes  \\
    \rowcolor{LightGray!50}
    & \vc{y}~\eqref{eq:ro-sub2} & yes & yes & yes  \\
    stereo loc. 
    & \color{LightRed} \vc{u}~\eqref{eq:u} & \color{LightRed}yes & \color{LightRed}no & \color{LightRed}no \\
    & \vc{u},$\vc{u}\kron\vc{t}$ & yes & yes & no \\
    \rowcolor{LightGray!50} 
    \ppr~\cite{briales_convex_2017} & none  & no & yes & yes \\
    \rowcolor{LightGray!50}
    \plr~\cite{briales_convex_2017} & none & yes & yes & yes \\
    \rppr~\cite{yang_certifiably_2022} 
    & \color{LightRed}$\vc{\theta}\kron\vc{\theta}$ & \color{LightRed}yes & \color{LightRed}no & \color{LightRed}no \\
    & $\vc{\theta}\kron\vc{w}$ & yes & yes & no \\
    \rowcolor{LightGray!50}
    \rplr~\cite{yang_certifiably_2022}
    & \color{LightRed}$ \vc{\theta}\kron\vc{\theta}$ & \color{LightRed}yes & \color{LightRed}no & \color{LightRed}no \\
    \rowcolor{LightGray!50}
    & $\vc{\theta}\kron\vc{w}$ & yes & yes & no \\
  \end{tabular}
\end{table}

\subsection{Hyperparameters}\label{sec:hyper}

Throughout the experiments, we keep the following parameters fixed. When learning the constraints, we oversample the data matrix $\vc{Y}$ by 20\% to improve conditioning of the nullspace problem. In  practice, we found this to be sufficient to ensure linearly independent constraints without significantly increasing the cost of the nullspace decomposition.
For the~\ac{SDP} solver, we use \texttt{MOSEK}~\cite{mosek} interfaced through \texttt{cvxpy}~\cite{cvxpy1,cvxpy2}, fixing the tolerances of primal and dual feasibility, as well as the relative complementary gap to $10^{-10}$ and the tolerance of infeasibility to $10^{-12}$. These tolerances were found to give satisfactory solution accuracy without impeding convergence. For finding the minimal set of constraints (Section~\ref{sec:sparsity}), we set the relative gap termination to $10^{-1}$ to allow even for inaccurate solutions to be returned (as the output is only used for ordering the constraints). 

In terms of local solvers, we use the off-the-shelf \texttt{pymanopt}~\cite{pymanopt} solver for all problems in Section~\ref{sec:results-other}, using the conjugate gradient optimizer and for stopping criteria $10^{-6}$ in gradient norm and $10^{-10}$ in step size. When inequality constraints are present in the~\ac{QCQP}, we use the log-sum-exp function described in~\cite[\S 4.1]{liu_simple_2020} with $\rho=10$ and $u=10^{-3}$. For~\ac{RO} localization, we use the \texttt{scipy} implementation of the~\texttt{BFGS} solver, and our custom~\ac{GN} implementation, respectively, with the same stopping criteria as for \texttt{pymanopt}. The above criteria were chosen to achieve good performance while attempting to yield similar accuracy as the SDP solver. 

In order to ensure the accuracy of the constraints, we compute the nullspace of $\vc{Y}$ twice, removing the sample from $\vc{Y}$ with highest nullspace errors after the first pass. This ensured that the maximal error was below $10^{-10}$ after the second pass for all problems.

As in~\cite{yang_certifiably_2022}, a problem is considered cost tight when its~\ac{RDG} is below $0.1\%$. It is considered rank tight when the~\ac{ER} is above $10^{7}$. Parameters that change for each problem, such as the considered noise levels, variable sets, and example problem sizes, are summarized in Table~\ref{tab:tight-problems}. The  chosen noise levels and problem sizes ensure unique and reasonably accurate solutions.. We use fully connected measurement graphs for all considered problems, as visualized in Figure~\ref{fig:setups}~(c).

\subsection{Range-Only Localization}\label{sec:results-range}

\textbf{\autotight:} We start by using \autotight~to evaluate the two different substitutions of \ac{RO} localization, on a small example problem as defined in Table~\ref{tab:tight-problems}.

\begin{figure}[t]
  \centering
  \begin{minipage}{.49\linewidth}\centering\footnotesize{$z_n$ substitution}\end{minipage}
  \begin{minipage}{.49\linewidth}\centering\footnotesize{$\vc{y}_n$ substitution}\end{minipage} \\
  \vspace{0.3em}
  \includegraphics[width=.48\linewidth]{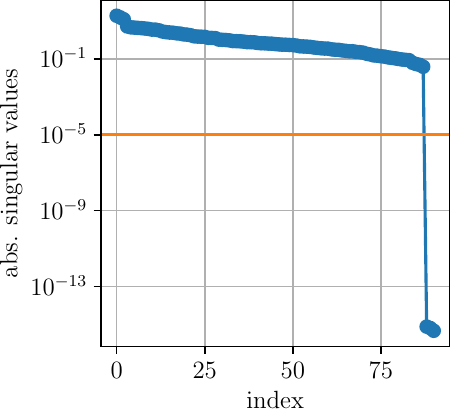}
  \includegraphics[width=.48\linewidth]{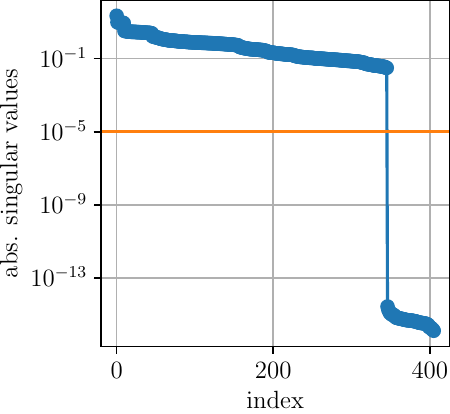}
  \caption{Singular value spectrum of the data matrix for~\ac{RO} localization. The singular values below the threshold (in orange) correspond to the nullspace basis vectors. For the substitution $z_n$~\eqref{eq:ro-sub1} (left plot), we find 3 basis vectors, however, for the substitution $\vc{y}_n$~\eqref{eq:ro-sub2} (right plot) we find 20 basis vectors.}\label{fig:svds}
\end{figure}

The data matrix $\vc{Y}$  exhibits a well-separated nullspace for both substitutions, as can be seen in Figure~\ref{fig:svds}. We can see immediately that the $z_n$ substitution leads to a small nullspace ($N_n=3=N$), corresponding exactly to the number of substitution variables. The substitution $\vc{y}_n$, on the other hand, leads to a nullspace that includes more than just the substitution variables ($N_n=60=20N$), which shows the existence of redundant constraints. 

\begin{figure}[t]
  \footnotesize{
    \includegraphics[width=\linewidth]{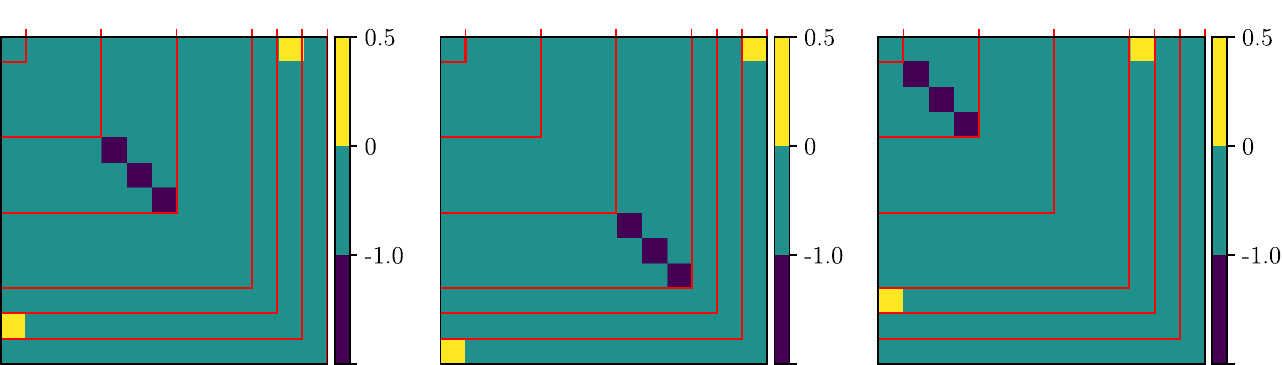}

    \vspace{0.1em}
    \begin{tabular}{C{2.0cm}C{0.2cm}C{2.0cm}C{0.2cm}C{2.0cm}}
    $z_2=\norm{\vc{\theta}_2}^2$  & 
    & $z_3=\norm{\vc{\theta}_3}^2$ & 
    & $z_1=\norm{\vc{\theta}_1}^2$ 
    \end{tabular}
    \includegraphics[width=\linewidth]{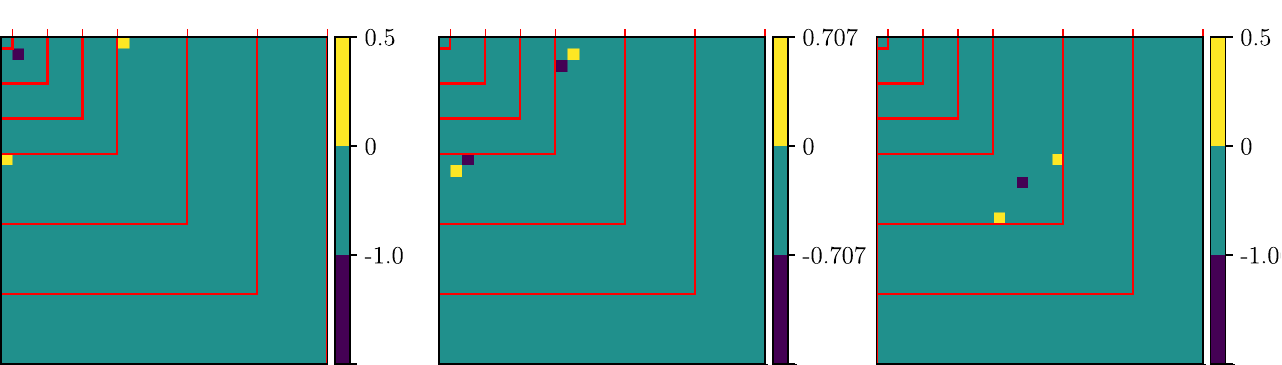}

    \vspace{0.1em}
    \begin{tabular}{C{2.0cm}C{0.2cm}C{2.0cm}R{2.8cm}}
    $\vc{y}[1]=a^2$   & 
    & $\vc{y}_1[1]b=\vc{y}_1[2]a$ $(a^2\cdot b=ab\cdot a)$   
    & $\vc{y}_1[1]\vc{y}_1[6]=\vc{y}_1[3]^2$ $(a^2\cdot c^2=(ac)^2)$
    \end{tabular}
  }
  \caption{Examples of learned constraint matrices for $z_n$ substitution (top) and the $\vc{y}_n$ substitution (bottom) of~\ac{RO} localization. Shown below each matrix are the algebraic identities that the matrices enforce. For simplicity, we call $\vc{\theta}_1^\top=\bmat{a & b & c}$.}\label{fig:ro-matrices}
\end{figure}

We show the three identified constraint matrices for the $z_n$ substitution in the first row of Figure~\ref{fig:ro-matrices}. Interestingly, the three automatically found matrices correspond exactly to the three substitution formulas (shown below each matrix).\footnote{Here, we chose not to enforce the known constraint matrices using~\eqref{eq:known}, to highlight the interpretability of the found constraints.} The second row of Figure~\ref{fig:ro-matrices} shows three example matrices for the $\vc{y}_n$ substitution. The first one is an example of a substitution constraint found by the algorithm, while the other two matrices are examples of discovered redundant constraints. Our method finds the $d(d+1)/2=6$ substitution constraints, and 14 redundant constraints similar to the two shown examples. 

We find that both substitutions lead to cost tight and rank tight relaxations when all constraints are imposed (including the redundant constraints for $\vc{y}_n$ substitution), with~\ac{ER} above $10^{9}$ and~\ac{RDG} below $10^{-4}$.  

\begin{table*}[t]
  \caption{Breakdown of characteristics for all tightened problems for the \last{formulation} phase of~\autoscale. This phase has to be run only once, and the output are reduced (red.) constraint (constr.) templates. All times are in seconds, with $t_n$ the total time to compute the nullspaces, $t_a$ the time to apply templates to all variables, $t_s$ the time to check for tightness, and $t_r$ the time required to reduce the constraints using~\eqref{eq:sparsity}.}\label{tab:times}
  \centering
  \begin{tabular}{l|l|cc|llll|l|cc}
  Problem & Dimension $n$ per iteration & \# Constraints & \# Reduced & $t_n$ [s] & $t_a$ [s] & $t_s$ [s] & $t_r$ [s] & total [s] & RDG & \pol{ER} \\ 
\midrule 
RO ($z_n$) & [10 15] & 4 & 4 & 0.01 &0.00 &0.17 &0.12 &0.29 & 4.41e-05 & \textbf{1.76e+09} \\ 
RO ($\vc{y}_n$) & [10 55] & 61 & 33 & 0.08 &0.09 &0.30 &9.25 &9.71 & 4.69e-05 & \textbf{2.27e+09} \\ 
stereo (2d) & [  28  168  546 1365] & 171 & 61 & 5.58 &0.92 &0.70 &13.32 &20.52 & \textbf{3.59e-06} & 4.39e+00 \\ 
stereo (3d) & [  91  910 3250 9100] & 639 & 225 & 299.35 &3.55 &3.30 &80.03 &386.23 & \textbf{2.86e-07} & 2.01e+01 \\ 
PPR & [91] & 21 & 7 & 0.21 &0.03 &0.17 &0.15 &0.56 & 3.31e-06 & \textbf{8.74e+09} \\ 
PPL & [91] & 21 & 10 & 0.21 &0.03 &0.40 &0.77 &1.40 & 1.85e-04 & \textbf{2.90e+08} \\ 
rPPR & [ 91 105 325 120 351 703] & 1771 & 1003 & 6.75 &12.94 &18.45 &548.16 &586.29 & \textbf{1.08e-06} & 3.44e+01 \\ 
rPLR & [ 91 105 325 120 351 703] & 2349 & 2049 & 6.79 &17.45 &22.14 &944.40 &990.78 & \textbf{4.05e-05} & 2.15e+01 \\ 

\end{tabular}
\end{table*}

\textbf{\secondmethod:} We use \autotemplate to find scalable templates for the $\vc{y}_n$ substitution, which requires a significant number of redundant constraints. In this particular example, the learned constraints are interpretable, as shown in Figure~\ref{fig:ro-matrices}, and we could infer the mathematical expression of all constraints (second outcome of Section~\ref{sec:tightness-summary}). Instead, we show here that \autotemplate is a tractable alternative that does not require any manual interpretation of constraint matrices. 

The employed variable sets are given in Table~\ref{tab:tight-problems}. The algorithm terminates after using group $\{\hom, \vc{\theta}_1, \vc{y}_1\}$, at which point the identified templates lead to a tight relaxation (in both cost and rank) when applied to all $N=3$ positions.\footnote{Note that we do not need to consider any combinations of positions (or substitutions), which is a consequence of the problem being separable. This could have been observed from~\eqref{eq:ro-original}, but we did not exploit this structure here to facilitate the extension to regularized problems (\ie, with motion prior).} 

\begin{figure}[t]
  \centering
  \begin{minipage}{.49\linewidth}\centering\footnotesize{$z_n$ substitution}\end{minipage}
  \begin{minipage}{.49\linewidth}\centering\footnotesize{$\vc{y}_n$ substitution}\end{minipage} \\
  \vspace{0.3em}
  \includegraphics[height=.43\linewidth]{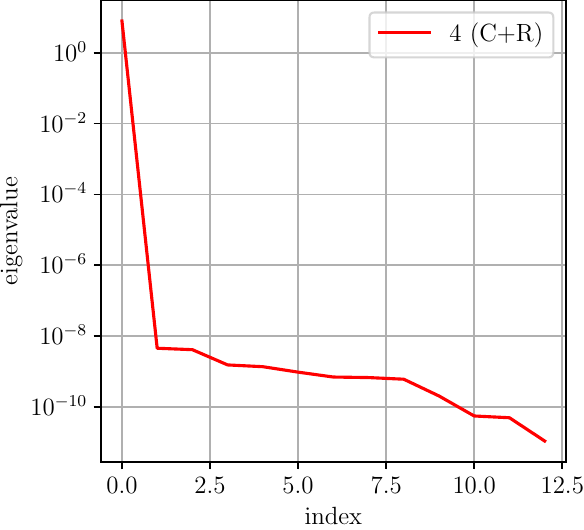}
  \includegraphics[height=.43\linewidth]{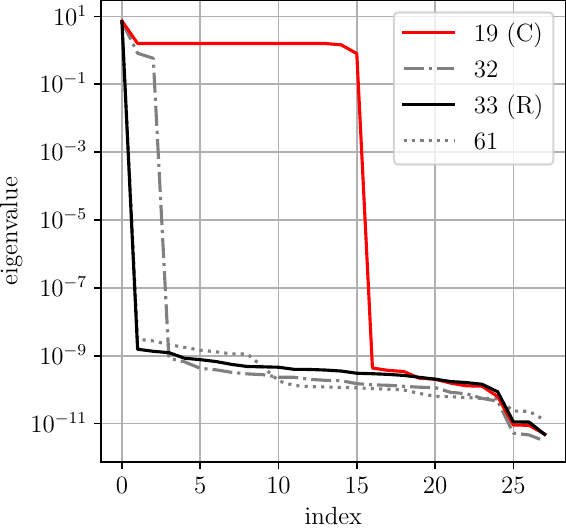}
  \caption{Rank-tightness study for \ac{RO} localization, using $z_n$ substitution (left) \vs~$\vc{y}_n$ substitution (right). We compare the spectra with different numbers of added constraints (gray lines), highlighting the points where cost tightness (C) and rank tightness (R) are obtained in red and black, respectively.}\label{fig:ro-eigs}
\end{figure}

Before applying the templates to new problems of increasing size, we reduce them to a smaller sufficient subset of constraints using~\eqref{eq:sparsity}. Figure~\ref{fig:ro-eigs} visualizes this process, showing rank- and cost tightness for different subsets of constraints used. First, we confirm that the substitution $z_n$ leads to rank- and cost tightness after adding the substitution constraints only. For $\vc{y}_n$, when adding constraints one-by-one in the order dictated by~\eqref{eq:sparsity}, we find that 33 out of the 61 constraints are enough for rank tightness. Cost-tightness, on the other hand, is achieved after adding 19 constraints only.

\begin{figure}[t]
  \centering
  \includegraphics[width=\linewidth]{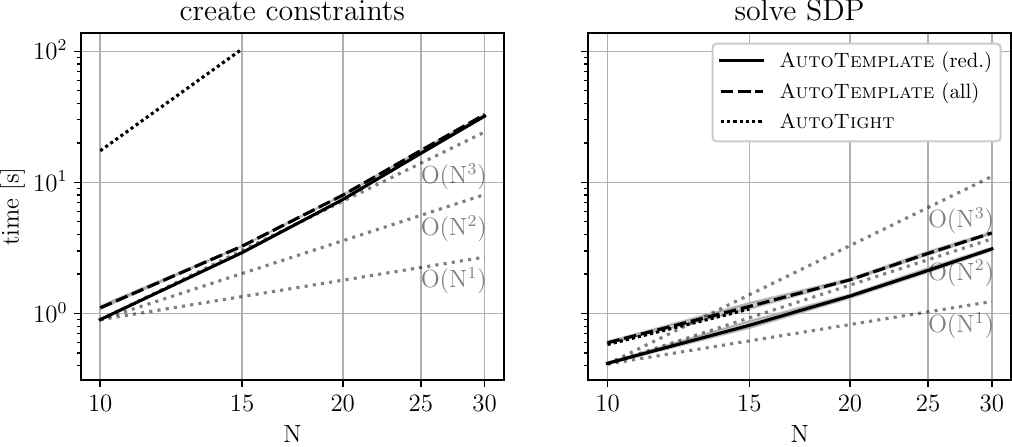}
  \caption{Timing study for~\ac{RO} localization, using the $\vc{y}_n$ substitution. We compare using only the reduced (solid line) or all templates (dashed line) output by~\autoscale, which are very close for this particular problem. They compare favorably to learning constraints from scratch for each problem using~\autotight.}\label{fig:results-time-ro}
\end{figure}

We apply the templates for $\vc{y}_n$ to problems with up to 30 positions. Figure~\ref{fig:results-time-ro} shows the time required for creating the constraints and solving the~\ac{SDP} for each problem size. We also report the results for $z_n$ in Appendix~\ref{app:redundant}. We compare the processing times \pol{of using \autotight from scratch \emph{vs.}} applying the templates from \autoscale, using either all or only the reduced constraint set sufficient for rank tightness.  

For the substitution $\vc{y}_n$\pol{,} \autotight becomes prohibitively expensive beyond $N=15$ positions. \pol{W}hen using \autoscale, the cost of generating the constraints \pol{becomes tractable, staying} close to the cost of solving the~\ac{SDP} for all problem sizes. Ordering the constraints according to~\eqref{eq:sparsity} did not have a significant effect in this case, and there is little difference between using the full \vs~only the reduced constraint set. Note that learning the templates and determining the reduced \last{constraints are part of the formulation phase and thus} constitute a fixed cost, listed separately in Table~\ref{tab:times}.

\subsection{Stereo Localization}\label{sec:results-stereo}

\textbf{\autotight:} We first use~\autotight to investigate whether the stereo localization problem~\eqref{eq:stereo-qcqp} can be tightened, using the example problem defined in Table~\ref{tab:tight-problems}. 

\begin{figure}[t]
  \centering
  \begin{minipage}{.49\linewidth}\centering\footnotesize{$\vc{u}_k$ substitution}\end{minipage}
  \begin{minipage}{.49\linewidth}\centering\footnotesize{$\vc{u}_k,\vc{u}_k\kron\vc{t}$ substitution}\end{minipage} \\
  \vspace{0.3em}
  \includegraphics[height=.42\linewidth]{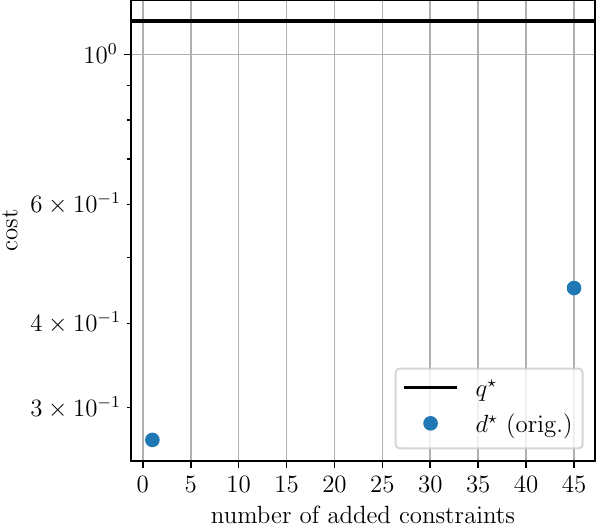}
  \includegraphics[height=.42\linewidth]{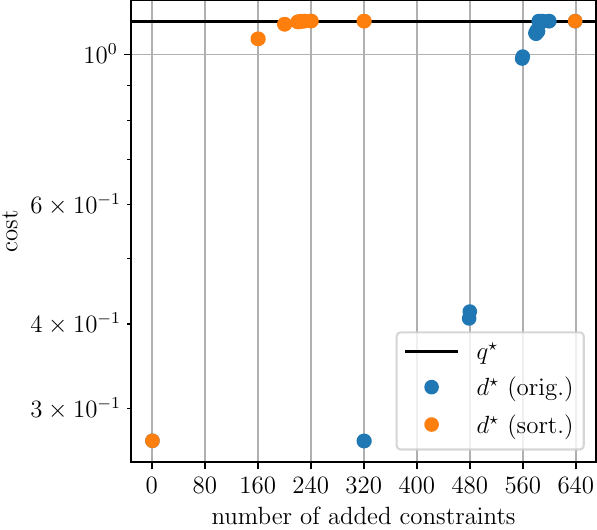}
  \caption{Tightness study for stereo localization, using the original substitutions (left) \vs~the higher-order substitutions (right). The bisection algorithm for finding the number of required constraints, terminates immediately for the original substitutions as even all constraints \pol{are not sufficient for cost tightness.} When adding higher-order substitutions, tightness is achieved after a few steps, using only 144 constraints when \pol{sorting constraints} using~\eqref{eq:sparsity} (sort.), and 590 when using the original order (orig.).}\label{fig:stereo2d3d-not-tight}
\end{figure}

The left plots of Figure~\ref{fig:stereo2d3d-not-tight} show the cost tightness study. Even when adding all 45 identified constraints, the problem cannot be tightened in the present form. Note how quickly we came to this conclusion: no manual search for redundant constraints had to be performed, a process that can be very time consuming. 

\begin{figure}[t]
  \begin{minipage}{.49\linewidth}\centering\footnotesize{original order}\end{minipage}
  \begin{minipage}{.49\linewidth}\centering\footnotesize{sorted order}\end{minipage} \\
  \vspace{0.3em}
  \includegraphics[height=.42\linewidth]{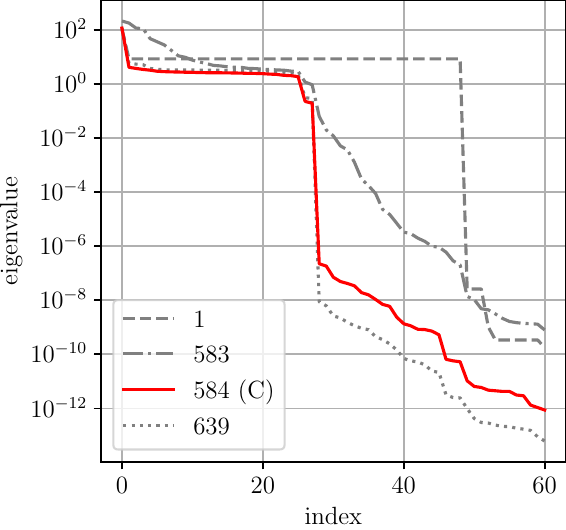}
  \includegraphics[height=.42\linewidth]{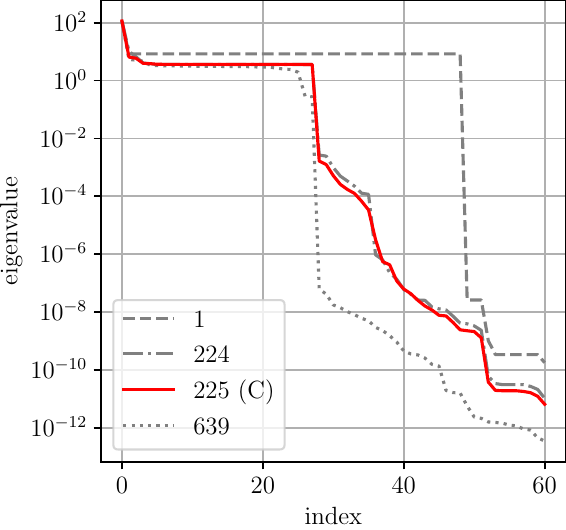}
  \caption{Study of the eigenvalue spectra of stereo localization using original order of constraints (left) and the sorted order using~\eqref{eq:sparsity} (right). Even after adding the higher-order substitutions and all redundant constraints, a significant number of eigenvalues are nonzero. More higher-order Lasserre variables may be required to achieve rank tightness. See Figure~\ref{fig:ro-eigs} for a detailed description of the labels.}\label{fig:stereo-eigs}
\end{figure}

We resort to (sparse) Lasserre hierarchy~\cite{lasserre} to tighten the problem. We try different higher-order lifting functions and retest for tightness after adding all possible redundant constraints. We individually test additions such as $\vc{u}_k \kron \vc{u}_k$, $\vc{t}\kron\vc{t}$, \etc~and find that by adding $(\vc{u}_k\kron \vc{t})$ for each landmark, we achieve tightness. For simplicity, we call the combined substitution $\vc{z}_k^\top:=\bmat{\vc{u}_k^\top & (\vc{u}_k\kron\vc{t})^\top}\inR{d+d^2}$. Figure~\ref{fig:stereo2d3d-not-tight} on the right shows the cost tightness test in 3D, which now passes. Since cost tightness is achieved, we can solve~\eqref{eq:sparsity} to determine a significantly smaller subset of sufficient constraints: we reduce the number from 639 to 144 constraints, as shown in Figure~\ref{fig:stereo2d3d-not-tight}. In all considered cases, rank tightness is not attained, which is shown in Figure~\ref{fig:stereo-eigs}. Rank tightness may require lifting functions of even higher order. As we are already approaching what is computationally feasible for the~\ac{SDP} solver, we settle for cost tightness.

\begin{figure}[t]
  \centering
  \footnotesize{
  \begin{tabular}{C{2.5cm}C{2.5cm}C{2.5cm}}
  \includegraphics[height=2.5cm]{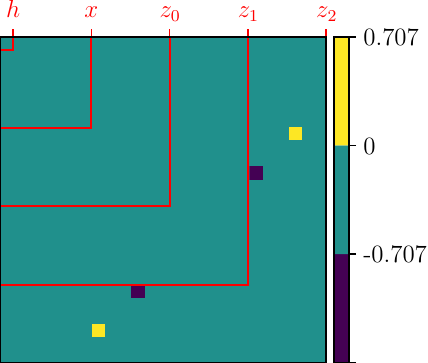}
  & \includegraphics[height=2.5cm]{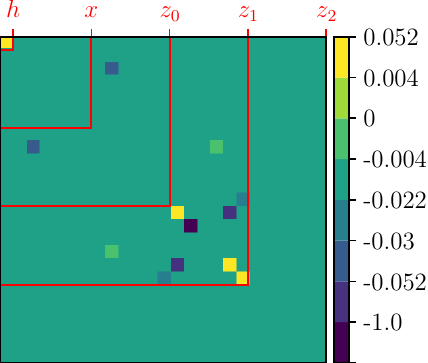}
  & \includegraphics[height=2.5cm]{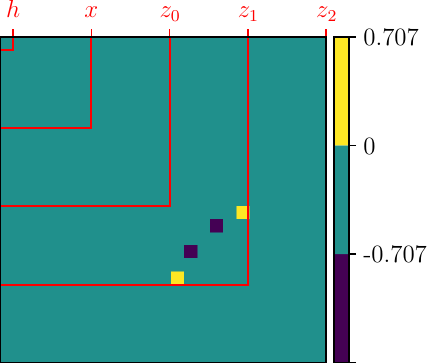} \\
  $ \vc{z}_1[1] \vc{z}_1[5] = \vc{z}_1[2]\vc{z}_1[3]$ $(\frac{1}{y}x{t}_x = \frac{1}{y} 1 \cdot t_xx)$ & not interpretable & not interpretable
  \end{tabular}
  }
  \caption{Three learned constraint matrices for a 2D stereo localization problem. Many of the matrices are less sparse than in the~\ac{RO} localization example and contain non-quantized numbers which suggests a dependency on landmark coordinates. Only few matrices, such as the one shown on the left, are interpretable (the identity is shown below the plot, where for simplicity, we call $\vc{t}={(t_x,t_y)}^\top$, $\vc{T}\vc{m}_2={(x, y, 1)}^\top$, thus $\vc{u}=\frac{1}{y}(x, 1)^\top$ and $\vc{z}_1=\frac{1}{y}{(x, 1, t_xx, {t}_yx, {t}_x, {t}_y)}^\top$).}\label{fig:stereo-matrices}
\end{figure}

\textbf{\autotemplate:} To tighten new problems, it is crucial to use \autoscale, for two reasons. Firstly, the problem dimension is large, in particular after adding the additional lifting functions required for tightness. Secondly, an investigation of the learned constraints, shown in Figure~\ref{fig:stereo-matrices}, suggests that many matrices actually depend on the (known) landmark coordinates and \pol{are therefore not easily interpretable}. 

\begin{figure}[t]
  \centering
  \includegraphics[width=\linewidth]{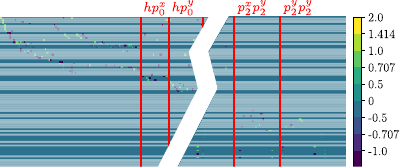}
  \caption{Subset of the constraint templates learned for stereo-localization in 2D after factoring out parameters. The red bars delimit different parameter dependencies, with the left-most block corresponding to the original variables. Highlighted in dark is the reduced set of templates sufficient for tightness (65 out of 171).}\label{fig:stereo2d-templates}
\end{figure}

Using the succession of variable sets listed in Table~\ref{tab:tight-problems}, we achieve tightness after including all groups up to $\{\hom, \vc{z_1}, \vc{z_2}\}$. Figure~\ref{fig:stereo2d-templates} shows the output of the method, for a 2D example: a set of templates over not only the original variables, but also their products with the parameters. 
Most importantly, note that \pol{the} matrix is now more quantized, with all nonzero elements in $\{2, \sqrt{2}, \pm1, \pm\frac{1}{\sqrt{2}}, \pm\frac{1}{2}\}$. We have thus \pol{factored out all} landmark dependencies and the obtained templates can be applied to any setup. The amount of templates may seem unmanageable at first; but the templates can be significantly reduced by solving~\eqref{eq:sparsity}: only 65 of the 171 templates (highlighted in dark in Figure~\ref{fig:stereo2d-templates}) are sufficient for tightness. 

\begin{figure}[t]
  \includegraphics[width=\linewidth]{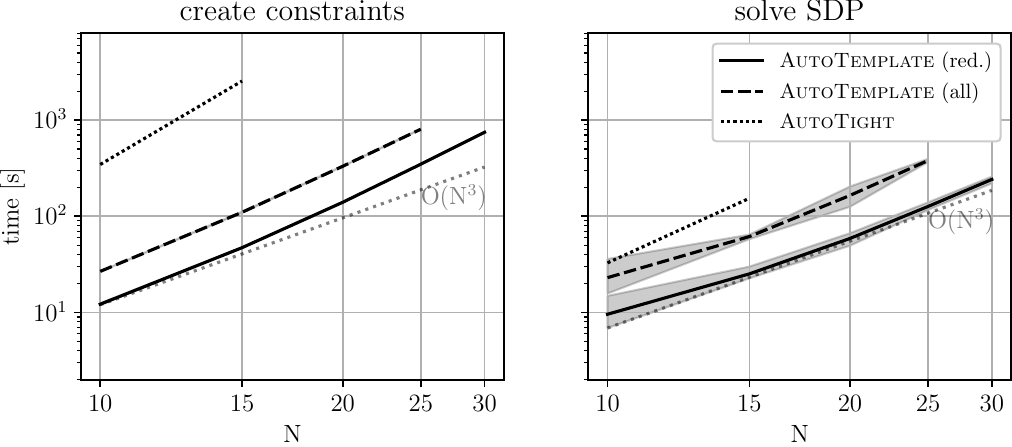}
  \caption{Timing study of the stereo-localization problem in 3D as we increase the number of landmarks $N$. The labels are the same as in Figure~\ref{fig:results-time-ro}. Learning constraints from scratch using~\autotight is prohibitively expensive even for $N=10$. On the other hand,~\autoscale scales reasonably up to $N=30$.}\label{fig:results-time-stereo}
\end{figure}

We successfully apply the templates for up to 30 landmarks. Figure~\ref{fig:results-time-stereo} shows how the times for \pol{applying the templates} and solving the~\ac{SDP} \last{(\ie, for the application phase)} scale with $N$\pol{. The} one-time cost for finding the \last{reduced set of templates (\ie, for the formulation phase) is} reported in Table~\ref{tab:times}. As for~\ac{RO} localization, learning templates from scratch for each new setup does not scale beyond $N=15$ landmarks, while applying the reduced templates comes at a reasonable cost, comparable to the cost of solving the~\ac{SDP} itself. This is a considerable improvement compared to existing approaches: inputting the same problem formulation to the (sparse) Lasserre hierarchy tool provided by~\cite{yang_certifiably_2022} leads to unmanageable numbers of variables and constraints, even for small problem sizes. For $d=3$ and only $N=3$ landmarks, a total of \num{27692} trivially satisfied constraints are generated, which is far beyond what~\ac{SDP} solvers can currently handle in reasonable time. In contrast, we can go to as many as $N=30$ landmarks, at which point we compute less than \num{5000} sufficient constraints for tightness.

\subsection{Other Problems}\label{sec:results-other}

We conclude the simulation study by applying the proposed method to a number of problems from the literature whose semidefinite relaxations have been shown to be tight using certain redundant constraints. We first consider two multimodal registration problems that have been treated by Briales~\etal\cite{briales_convex_2017}: point-point registration (\ppr) and point-line registration (\plr), before studying their robust version, provided by Yang~\etal\cite{yang_certifiably_2022}, in the next section.

\subsubsection{\ppr and \plr~\cite{briales_convex_2017}} In multimodal registration, the goal is to find an object's translation $\vc{t}\inR{d}$ and orientation $\vc{C}\in\SO{d}$ \wrt~a world frame, given measurements of points lying on the object. The object is assumed to be represented by a set of known geometric primitives of either points (\ppr), lines (\plr), or planes (not considered for brevity). The problem is posed as the following minimization problem~\cite{briales_convex_2017}:
\begin{equation}
  \min_{\vc{C}\in\SO{d}, \vc{t}\inR{d}} \sum_{i=1}^{N} \norm{\vc{C}\vc{p}_i +\vc{t}-\vc{y}_{i} }^2_{\vc{W}_i},
  \label{eq:briales}
\end{equation}
with $\vc{p}_i\inR{d}$  the measured point and $\vc{y}_i$ an arbitrary point on the associated primitive $P_i$ (note that data association is assumed known). The matrix $\vc{W}_i\succ 0\inR{d\times d}$ is chosen depending on the type of primitive $P_i$, see Appendix~\ref{app:math}.

Problem~\eqref{eq:briales} can be relaxed to a~\ac{QCQP} by dropping the determinant constraint from $\SO{d}$ as explained in Section~\ref{sec:results-stereo}, and introducing $\vc{x}^\top=\big[\hom \,\, \vc{\theta}^\top\big]$, with $\vc{\theta}^\top=\big[\vc{t}^\top \, \ve{(\vc{C})}^\top\big]$.

\textbf{Manual method~\cite{briales_convex_2017}:} The primal relaxation of problem~\eqref{eq:briales} was shown to be always tight when using a specific set of redundant constraints, that enforce, for example, the handedness of the $\vc{C}$ matrix that may have been lost because of dropping the determinant constraint~\cite{briales_convex_2017}. The formulas of these 22 redundant constraints are given in Appendix~\ref{app:math}.

\textbf{Proposed method:} \autotight finds the required redundant constraints outlined above automatically. As shown in Appendix~\ref{app:redundant}, we find a total of 21 independent constraints, including the homogenization, suggesting that at least one of the 22 constraints presented by~\cite{briales_convex_2017} \pol{is} linearly dependent. Indeed \pol{let $e_i(\vc{C})=1$ be the orthonormality constraints involving} the diagonal with $i\in\{1, 2, 3\}$ for~\eqref{eq:o1} and $i\in\{4, 5, 6\}$ for~\eqref{eq:o2}. 
Then, it is easy to see \pol{that} $\sum_{i=1}^3 e_i(\vc{C}) = \sum_{i=4}^6 e_i(\vc{C})$,
so any of these six constraints can be written as a linear combination of the five others.

\begin{figure}[t]
  \centering
  \begin{minipage}{.49\linewidth}\centering\footnotesize{\ppr}\end{minipage}
  \begin{minipage}{.49\linewidth}\centering\footnotesize{\plr}\end{minipage} \\
  \vspace{0.3em}

  \includegraphics[width=.49\linewidth]{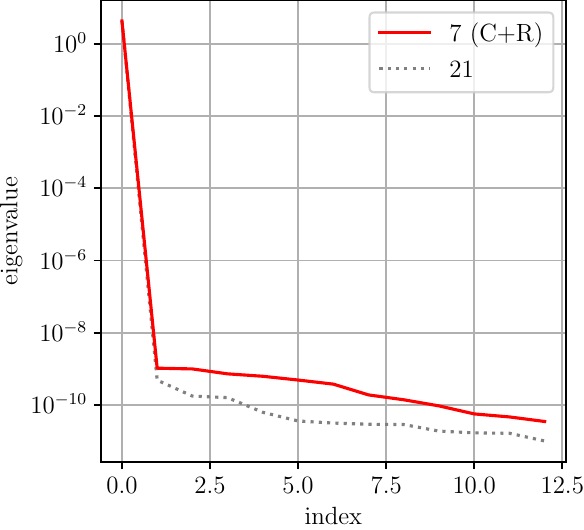}
  \includegraphics[width=.49\linewidth]{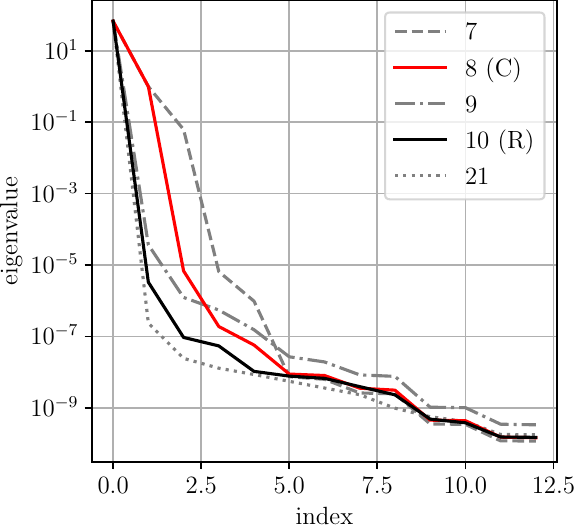}
  \caption{Rank-tightness study for \ppr~(left) and \plr~(right) problems~\cite{briales_convex_2017}. \ppr is cost tight without redundant constraints, and for \plr, only one and three redundant constraints are required for cost and rank tightness, respectively; a small subset of the 12 available redundant constraints~\cite{briales_convex_2017}.}\label{fig:briales-eigs}
\end{figure}

While the constraints by~\cite{briales_convex_2017} have been shown to be \textit{sufficient} for tightness, they have not been shown to be \textit{necessary}. In fact, we found that, for the considered noise level, none of the redundant constraints are required for \ppr~to be both cost and rank tight, as shown in Figure~\ref{fig:briales-eigs}. For \plr, Figure~\ref{fig:briales-eigs} shows that the solution becomes rank one after adding as few as three of the 12 available redundant constraints. 

\subsubsection{rPPR and rPLR~\cite{yang_certifiably_2022}\label{short47}} 

Next, we consider the robust versions of the two registration problems: \rppr and \rplr. These problems are called robust pointcloud registration and robust absolute-pose estimation, respectively, in~\cite{yang_certifiably_2022}, and they are two (of many) robust estimation problems that can be formulated as a QCQP. Using the~\ac{TLS} cost function as an example, \rppr and \rplr can be formulated as
\begin{equation}
  \begin{aligned}
    \min_{\vc{\theta}\in\mathcal{D}, \vc{w}\in\{\pm 1\}^N} & \frac{1}{2}\sum_{i=1}^N \frac{1 + w_i}{\beta_i^2}r^2(\vc{\theta},\vc{y}_i) + {1 - w_i},
\end{aligned}
  \label{eq:robust-tls}
\end{equation}
where $\vc{y}_i$ are measurements, $\vc{w}$ is the vector of decision variables (for outliers, $w_i=-1$ and for inliers $w_i=1$) and $\beta_i>0$ are user-defined parameters determining the truncation threshold. The feasible domain $\mathcal{D}$ and a sketch of the derivation of \eqref{eq:robust-tls} are given in Appendix~\ref{app:math}. The residual functions are given by 
\begin{align}
  \text{\rppr:} \quad r(\vc{\theta}, \vc{y}_i)&=\norm{\vc{C}\vc{p}_i + \vc{t} - \vc{y}_i}^2,  \\
  \text{\rplr:} \quad r(\vc{\theta}, \vc{y}_i)&= \norm{\vc{C}\vc{p}_i + \vc{t}}_{\vc{I}_d-\vc{y}_i\vc{y}_i^\top}^2,
\end{align}
where $\vc{y}_i$ are pointcloud and unit direction measurements, respectively, in \rppr and \rplr.
Problem~\eqref{eq:robust-tls} can be written as a~\ac{QCQP} in the lifted vector
\begin{equation}
  \vc{x}^\top = \bmat{\hom & \vc{\theta}^\top & \vc{w}^\top & \vc{z}^\top},
  \label{eq:x-robust}
\end{equation}
with $\vc{\theta}^\top = [\vc{t}^\top \, \vectorised{\vc{C}}^\top]$. The variable $\vc{z}$ contains additional substitutions that are required to make problem~\eqref{eq:robust-tls} quadratic in $\vc{x}$ (the cost is cubic because the residual functions $r$ are linear in $\vc{\theta}$), as discussed next.

\textbf{Manual method~\cite{yang_certifiably_2022}:} In~\cite{yang_certifiably_2022}, the authors propose to add the (sparse) Lasserre lifting function  $\vc{z}=\vc{\theta}\kron\vc{w}$, which leads to a tight relaxation after adding a list of (trivially satisfied) constraints. The authors also mention in passing that other lifting functions, such as $\vc{z}=\vc{\theta}\kron\vc{\theta}$, which also allow to write~\eqref{eq:robust-tls} as a~\ac{QCQP}, do not lead to a tight relaxation. 

\textbf{Proposed method:} We study both lifting functions and come to the same conclusions as in~\cite{yang_certifiably_2022}: both formulations allow for a large number of redundant constraints (which we find automatically), but only the first formulation becomes tight. Because of the large number of variables in the lifted state vector, we resort directly to \autoscale. The variable ordering used (for both problems) can be found in Table~\ref{tab:tight-problems}, where we drop some of the trivial first variable sets to save time. When using the lifting function $\vc{z}:=\vc{\theta} \kron \vc{w}$, the method terminates with cost tightness after considering variables $\{l, \vc{\theta}, \vc{z}_1, \vc{z}_2\}$. For $\vc{z}:=\vc{\theta}\kron\vc{\theta}$, the method returns that no tightness can be achieved. 

\begin{figure}[t]
  \centering
  \includegraphics[width=\linewidth]{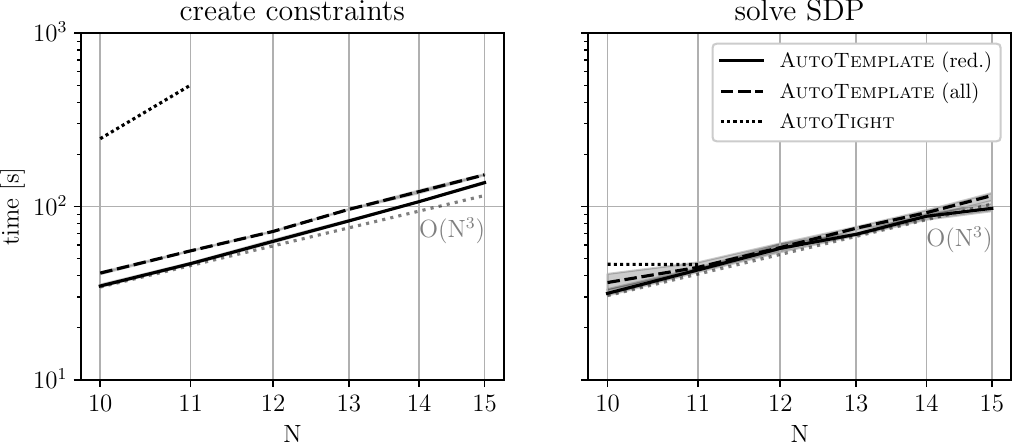}
  \caption{Timing results of scaling to $N$ landmarks for \rppr. Thanks to \autoscale, we can automatically create the constraints of problems up to $N=15$ landmarks.}\label{fig:robust-autotemplate}
\end{figure}

The number of constraint templates before and after reduction can be found in Table~\ref{tab:times}. \pol{The} number of required constraints is already very high (more than \num{1000}) when considering only $N=4$ and $N=6$ for \rppr and \rplr, respectively. Nevertheless, we can apply the templates to problems up to size $N=15$, as shown in Figure~\ref{fig:robust-autotemplate}, for \rppr. We report the timing results for \rplr, and the eigenvalue spectra for \rppr and \rplr, in Appendix~\ref{app:redundant}. For both problems, learning constraints from scratch is prohibitively expensive. \pol{With \autoscale,} we obtain cost tightness for all considered problems. Just as in stereo localization, rank tightness is not achieved, as shown in Figure~\ref{fig:robust-eigs} in the Appendix, and is not computationally tractable since we already need many constraints for cost tightness. 

As a final study, we compare the number of constraints we find with the number of constraints found in~\cite{yang_certifiably_2022} in Table~\ref{tab:comparison}. The results suggest that we find a significantly smaller subset of constraints, but without compromising tightness. One possible explanation is that we find more than only the ``trivially satisfied'' redundant constraints at each level, and thus we can chose from a larger pool when tightening the problem. We plan to further investigate this \pol{phenomenon in future research}. 

\begin{table}[ht]
  \vspace{1em}
  \caption{The number of constraints for cost tightness found for \rppr~and \rplr, respectively, using our method and the method proposed by~\cite{yang_certifiably_2022}, as a function of the number of measurements $N$. }\label{tab:comparison}
  \centering
  \begin{tabular}{cccccc}
  N & \multicolumn{2}{c}{\rppr} & \multicolumn{2}{c}{\rplr} & \\
  \midrule
  & our method &\cite{yang_certifiably_2022} & our method &\cite{yang_certifiably_2022} \\
  \midrule
  10& 4,508& 6,257 &  5,330 & 7,379 \\
  11& 5,293& 7,398 &  6,279 & 8,724 \\
  12& 6,139& 8,633 &  7,304 & 10,180 \\
  13& 7,046& 9,962 &  8,405 & 11,747 \\
  14& 8,014& 11,385&  9,582 & 13,425 \\
  15& 9,043& 12,902&  10,835& 15,214\\
\end{tabular}
\end{table}

\section{Real-world Experiments}\label{sec:results-real}

\begin{figure*}[t]
    \centering
    \begin{minipage}[t]{.62\linewidth}
      \vspace{0pt}
      \centering
      \includegraphics[width=.49\linewidth,trim={0 0 2cm 0},clip]{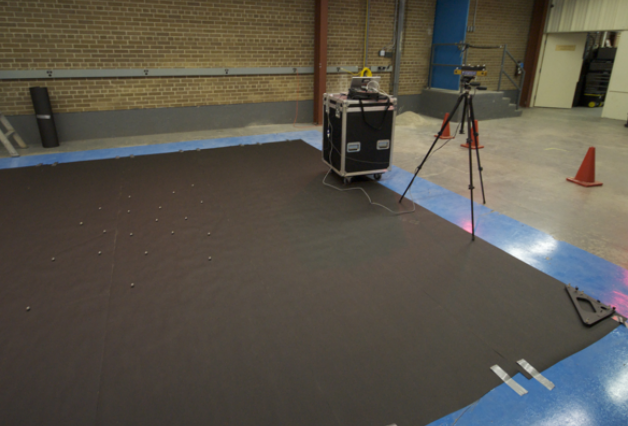}%
      \llap{
        \includegraphics[height=2cm]{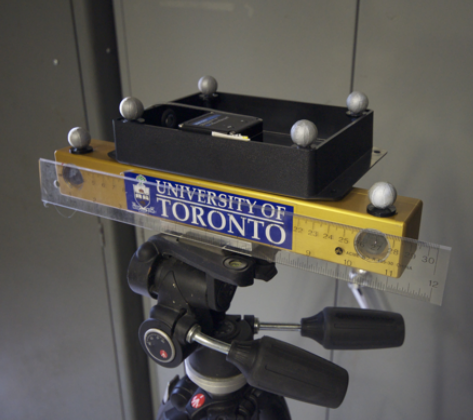}%
        \hspace{0.05cm}
      }%
      \llap{\raisebox{2.1cm}{
          \includegraphics[width=.17\linewidth]{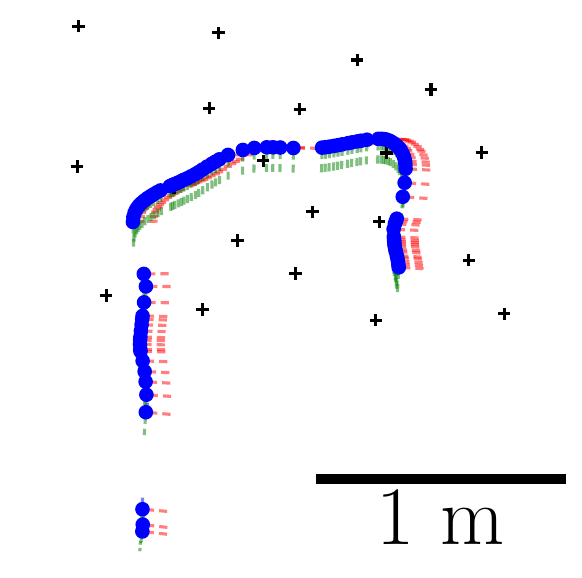}%
          \hspace{0.05cm}
      }}
      \centering
      \includegraphics[width=.49\linewidth]{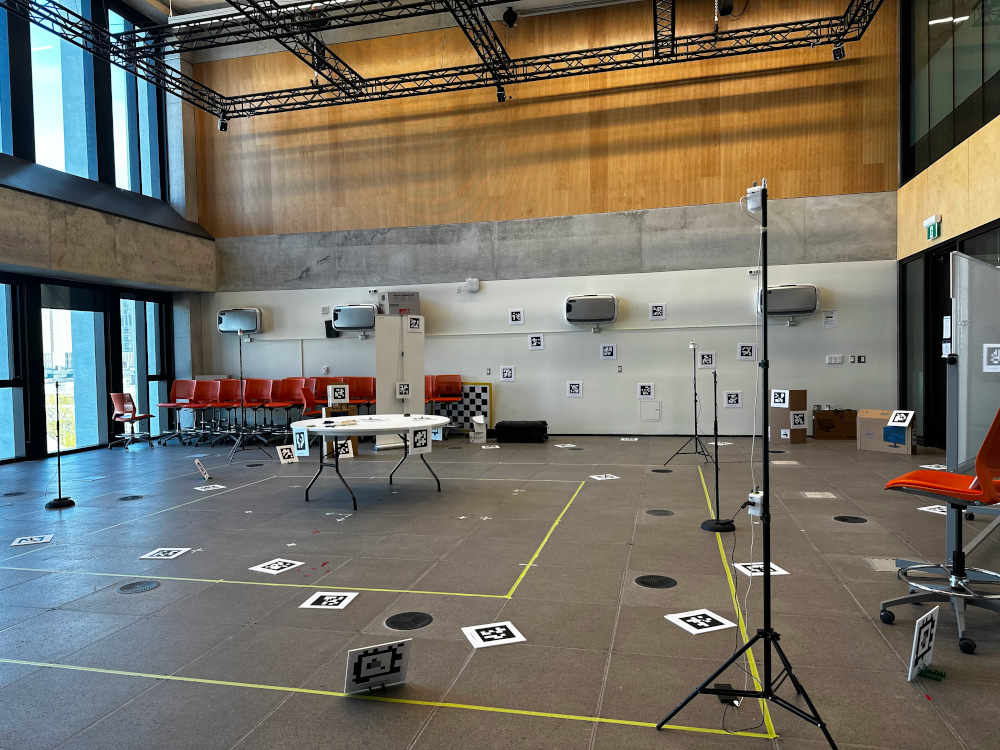}%
      \llap{\raisebox{0.0cm}{
          \includegraphics[height=2cm,trim={0cm 0cm 0 4cm},clip]{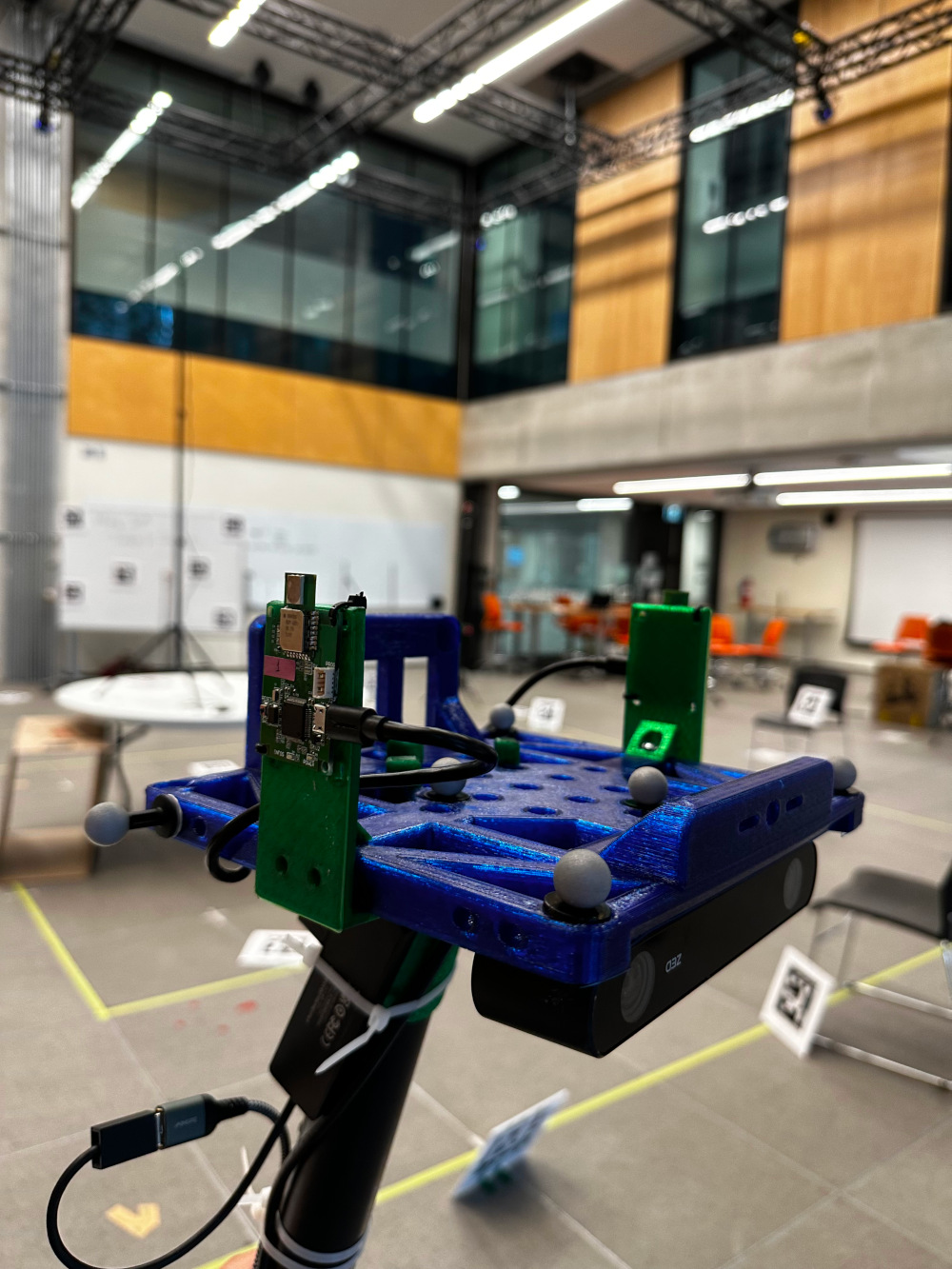}%
          \hspace{0.05cm}
      }}
    \end{minipage}
    \begin{minipage}[t]{.37\linewidth}
      \vspace{0pt}
      \includegraphics[width=.32\linewidth]{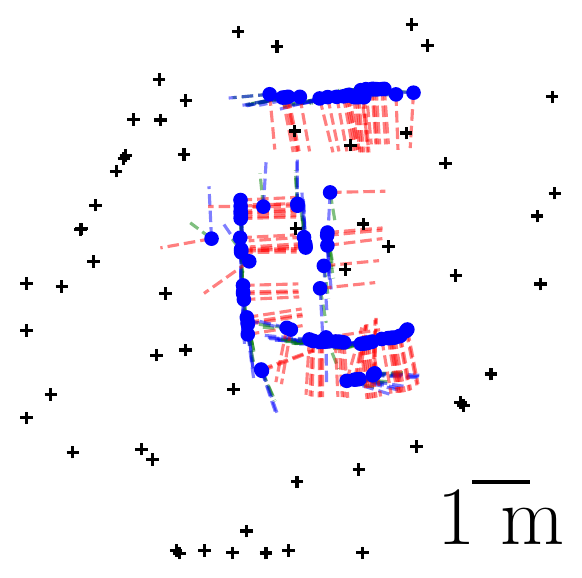}%
      \includegraphics[width=.32\linewidth]{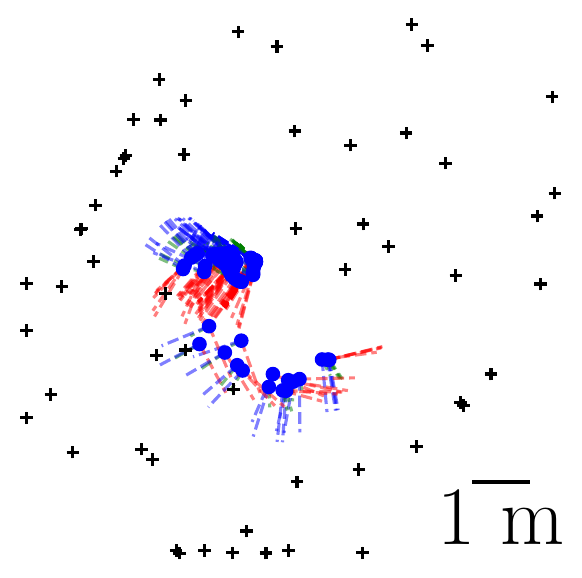}%
      \includegraphics[width=.32\linewidth]{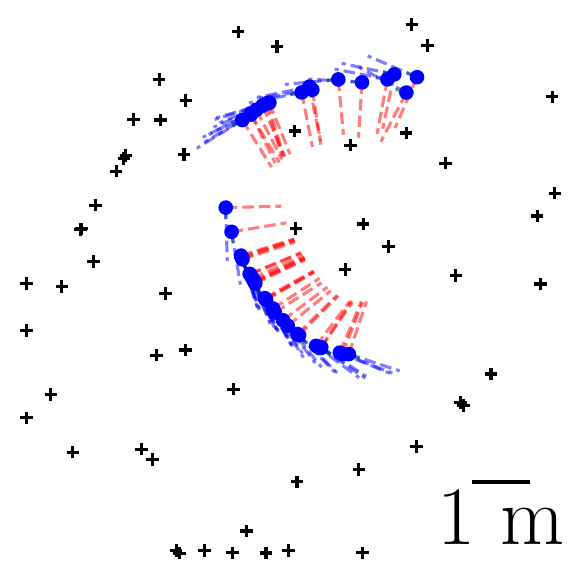}\\
      \includegraphics[width=.32\linewidth]{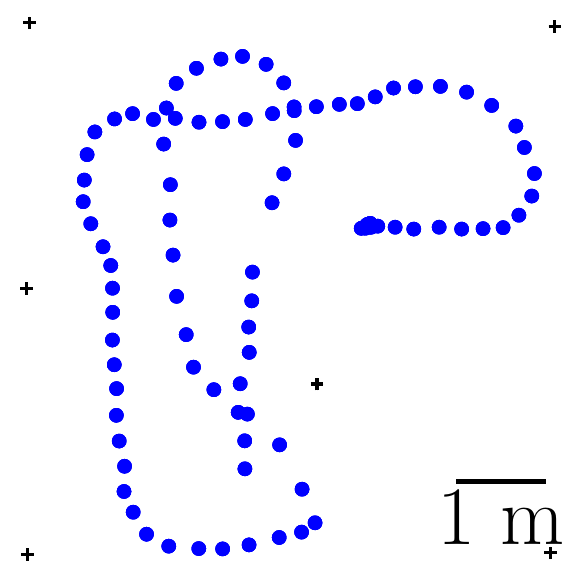}%
      \includegraphics[width=.32\linewidth]{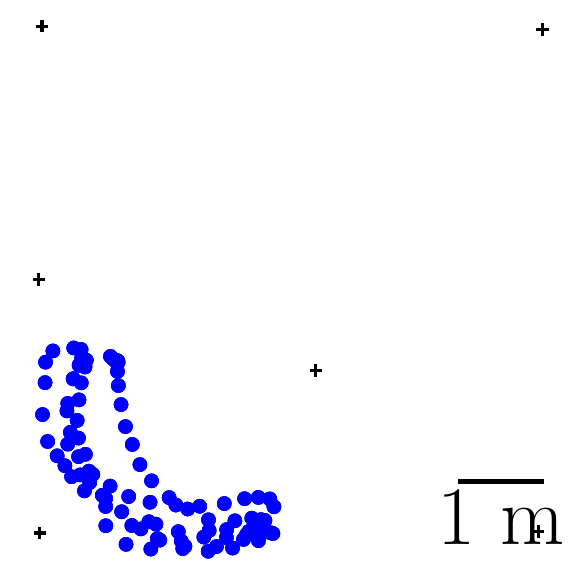}%
      \includegraphics[width=.32\linewidth]{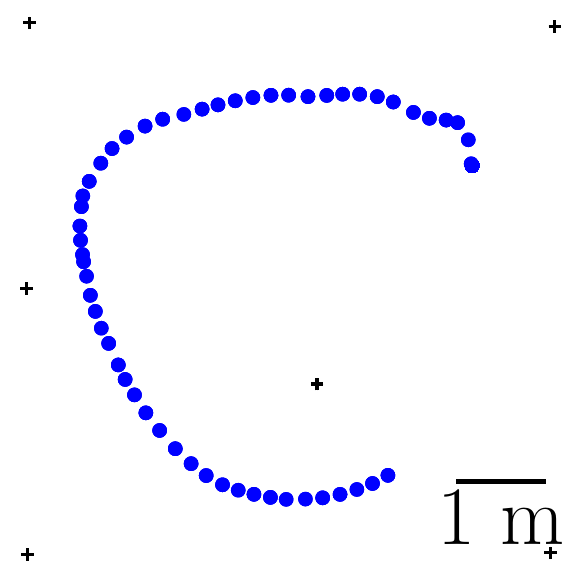}
    \end{minipage}
    \begin{minipage}[t]{.31\linewidth}\centering\footnotesize{\textbf{\textit{starrynight}}}\end{minipage}
    \begin{minipage}[t]{.31\linewidth}\centering\footnotesize{\textbf{\textit{STAR-loc}}}\end{minipage}
      \begin{minipage}[t]{.11\linewidth}\centering\footnotesize{\textit{zigzag\_s3}}\end{minipage}
      \begin{minipage}[t]{.11\linewidth}\centering\footnotesize{\textit{eight\_s3}}\end{minipage}
      \begin{minipage}[t]{.11\linewidth}\centering\footnotesize{\textit{loop-2d\_s4}}\end{minipage}
      \caption{\pol{Experimental setups of real-world datasets: \textit{starrynight} (left)~\cite{starrynight}, and \textit{STAR-loc} (middle \& right plots)~\cite{starloc}. \textit{Apriltags} The plots show the ground-truth poses at which stereo (top) and \ac{UWB} (bottom) measurements from fixed landmarks (black crosses) are obtained.}}\label{fig:datasets}
\end{figure*}

To conclude, we showcase the performance of the proposed framework on real-world datasets for~\ac{RO} localization and stereo localization. The purpose of these experiments is to
1) give an example of the full pipeline in action \pol{and} to
2) investigate how the constraints, determined using a specific example problem, generalize to real data with different characteristics. \last{In the formulation phase, we} run \autoscale~with random landmarks to generate constraint templates. No knowledge of the actual measurement setup is required at this point. \last{In the application phase, we} \pol{apply} the templates to generate constraints, \pol{using} the actual landmark locations at each considered pose. 

\subsection{Experimental Setups}\label{sec:datasets}

We test our methods on two \pol{real-world datasets}\pol{, visualized in Figure~\ref{fig:datasets}. The \textit{starrynight} dataset~\cite{starrynight}} includes stereo-camera images of Vicon markers scattered on the floor. \pol{The \textit{STAR-loc} dataset~\cite{starloc}} includes stereo-camera images of \textit{Apriltag}~\cite{apriltags} landmarks scattered around a room at different heights and orientations. The \textit{STAR-loc} dataset also includes \ac{UWB}-based distance measurements to eight fixed anchors. 

For~\ac{RO} localization, we always randomly select 4 out of the 8 available anchors to investigate the local minima that typically arise when anchors are almost co-planar. We report results on three example runs: \textit{zigzag\_s3}, \textit{loop-2d\_s4}, and \textit{eight\_s3}. For stereo localization, we only consider poses where more than 4 landmarks are observed, and we cap at maximum 8 landmarks, to limit the computation time. \last{We initialize the local solver with normally distributed positions centered at the ground truth and with \SI{1}{m} standard deviation in all axes.}

\begin{figure}[tb]
  \centering
  \footnotesize{
  \begin{minipage}{.49\linewidth} \centering \ac{RO} localization \end{minipage}
  \begin{minipage}{.49\linewidth} \centering stereo localization \end{minipage} \\
  }
  \vspace{0.3em}
  \includegraphics[width=.49\linewidth]{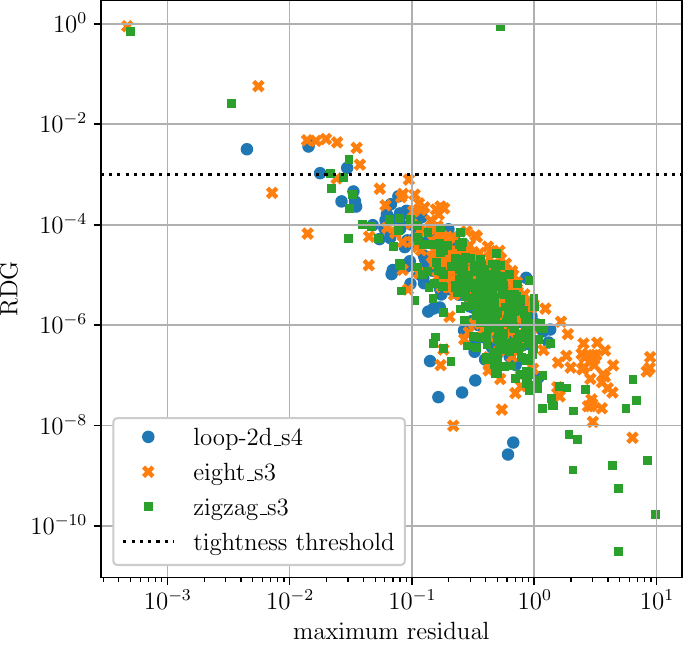}
  \includegraphics[width=.49\linewidth]{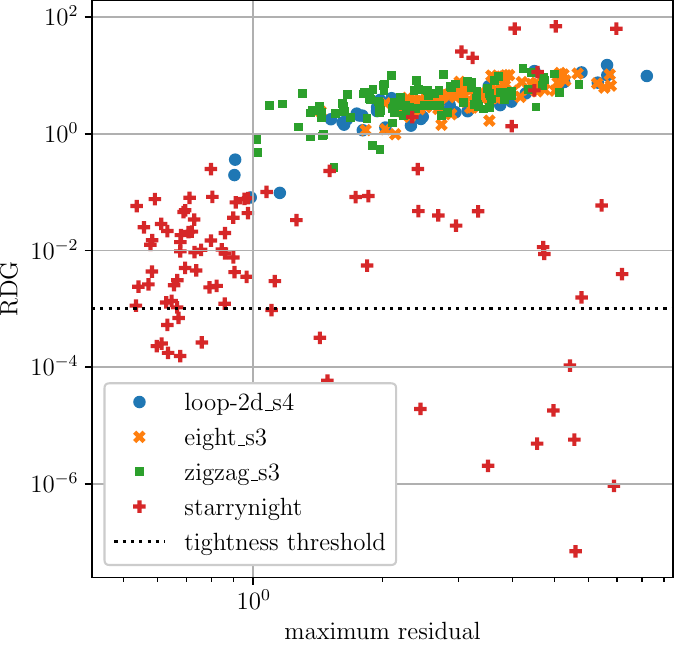}
  \includegraphics[width=.49\linewidth]{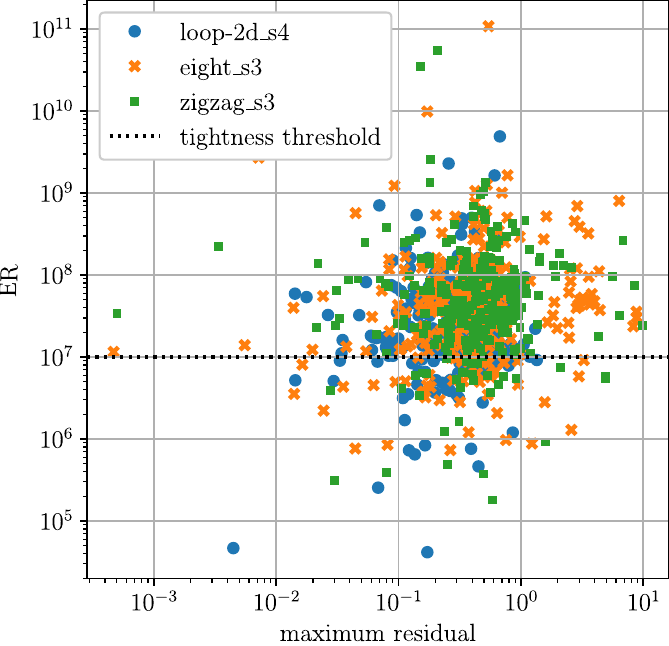}
  \includegraphics[width=.49\linewidth]{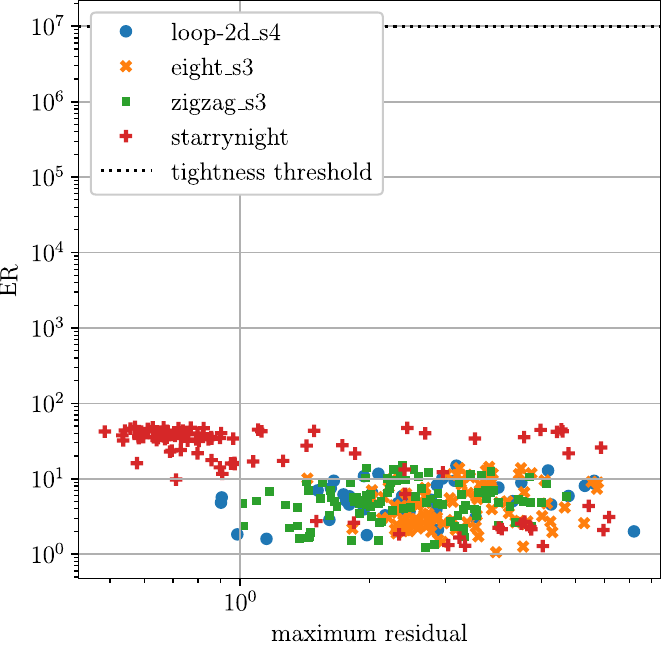}
  \caption{\pol{Real-world} tightness study for~\ac{RO} (left plots) and stereo localization (right plots). Each data point corresponds to one estimated pose. Cost-tightness (top) and rank tightness (bottom) are compared with the maximum residual \pol{error.} We see that \ac{RO} localization is mostly cost- and rank tight, while stereo localization is only cost tight for the lowest residual error levels in the \textit{starrynight} dataset.}\label{fig:real-tightness}
\end{figure}

\subsection{Results}

First, we investigate the tightness of the relaxations when evaluated on real data. Figure~\ref{fig:real-tightness} shows the~\ac{ER} and~\ac{RDG} for both \ac{RO} and stereo localization, for randomly picked poses from both datasets (see Figure~\ref{fig:datasets} for plots of the selected poses). We plot the respective tightness measures against the maximum residual error, which is a good proxy for the noise level and has been shown to affect the tightness of semidefinite relaxations~\cite{eriksson_rotation_2018,cifuentes_local_2022}. As expected, the relaxation of~\ac{RO} localization is mostly rank tight across all considered datasets and poses, with an~\ac{ER} of more than $10^6$ for most poses. On the other hand, the stereo-localization relaxation is only reliably cost tight for poses with a sub-pixel maximum residual error, which is a characteristic found in the \textit{starrynight} dataset but in none of the runs from the~\textit{STAR-loc} dataset.

\begin{figure}[tb]
  \centering
  \footnotesize{
  \begin{minipage}{.49\linewidth} \centering \ac{RO} localization \end{minipage}
  \begin{minipage}{.49\linewidth} \centering stereo localization \end{minipage}\\
  }
  \vspace{0.3em}
  \includegraphics[width=.49\linewidth]{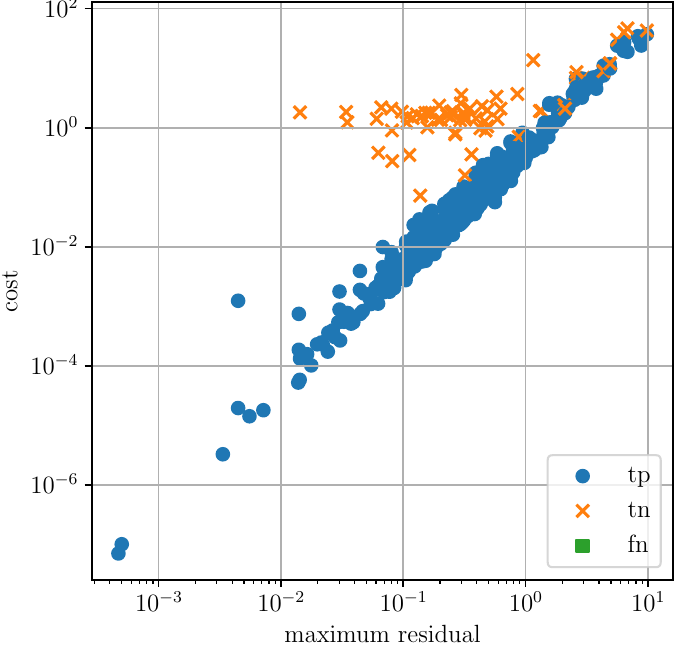}
  \includegraphics[width=.49\linewidth]{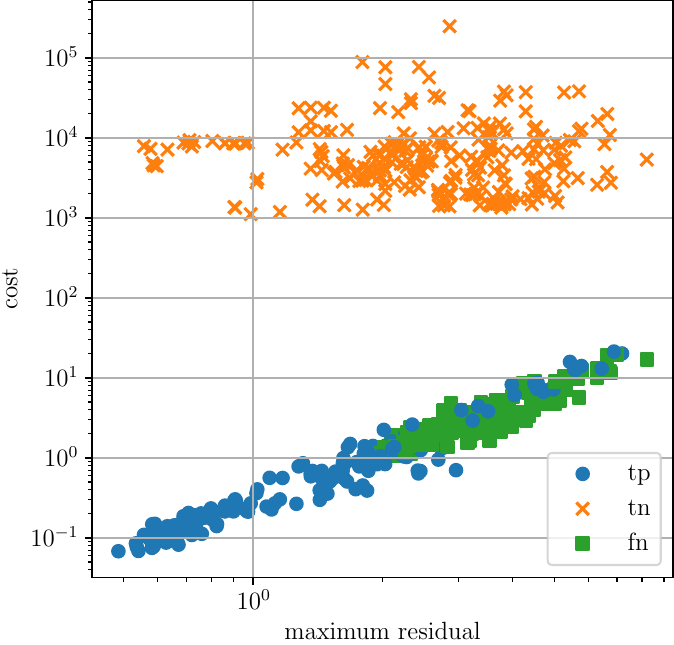}
  \caption{\pol{Real-world} comparison \pol{of} local and global solutions as a function of the maximum residual error. Each data point corresponds to one pose \pol{estimate}, and we distinguish between true positives (\emph{tp}, certified global minimum), true negatives (\emph{tn}, uncertified local minimum), false negatives (\emph{fn}, uncertified global minimum). Crucially, no false positives occurred. Note that all local solutions for stereo localization exhibit a high cost, and as the maximum residual error increases, \pol{false negatives occur}, which is a consequence of \pol{reduced tightness } (compare with Figure~\ref{fig:real-tightness}).}
  \label{fig:real-local-error}
\end{figure}

Next, we study the occurrence of local \vs~global minima found in both problems. We certify a local solution by trying to find dual variables that satisfy~\eqref{eq:certificate}  via a feasibility~\ac{SDP}. To account for numerical errors, we change~\eqref{eq:stationarity} to $|\vc{H}(\rho, \vc{\lambda})|\leq\epsilon\vc{1}$ and minimize $\epsilon$ as objective function. We claim that a candidate solution $\hc{x}$ is certified if we find a feasible solution with $\epsilon \leq 10^{-3}$.  

Figure~\ref{fig:real-local-error} shows the \pol{cost} of certified and uncertified solutions of~\ac{RO} and stereo localization, as a function of the maximum residual error. \pol{For \ac{RO} localization}, we note that local minima are ubiquitous for both problems, across all noise levels. Second, it can be observed that, as the noise increases, the global solution candidates from the stereo-localization problem are not all certified anymore, because of the lack of tightness at higher noise levels. However, local minima occur even at lower levels, and the relaxation can correctly identify them. For \ac{RO} localization, all global solutions are correctly certified. All local solutions, which, compared to stereo localization, are harder to classify based on only their cost, are \pol{also} detected correctly. Note that since this relaxation is rank tight, solving the primal~\ac{SDP} and extracting $\vc{x}^\star$ from the rank-1 $\vc{X}^\star$ would also be a viable solution method.

\begin{figure}[t]
  \begin{minipage}{.22\linewidth} \centering 
    \includegraphics[height=3.0cm]{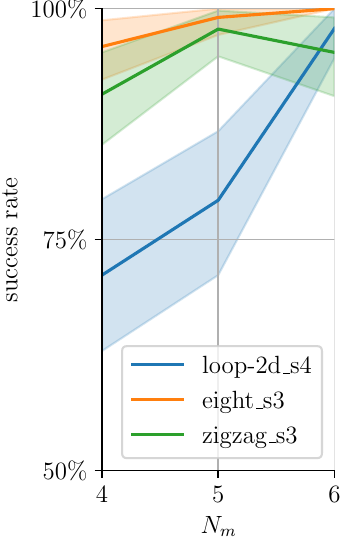}
  \end{minipage}
  \begin{minipage}{.38\linewidth} \centering \footnotesize{ \textit{loop-2d\_s4},  $N_m=4$} \\
  \includegraphics[trim={0 0cm 0 1cm},clip,width=\linewidth]{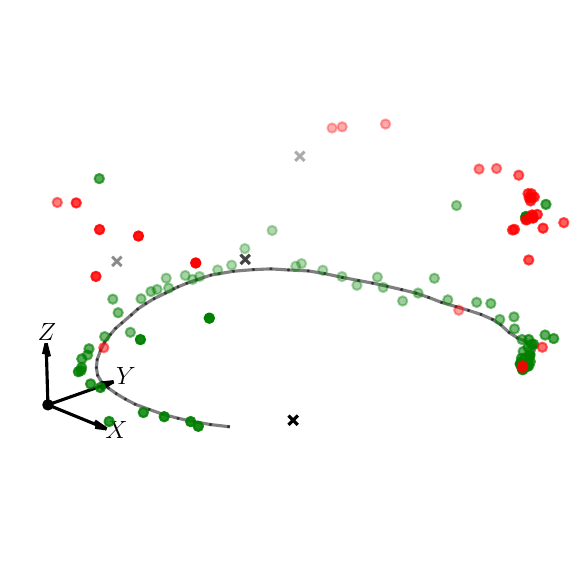}
  \end{minipage}
  \begin{minipage}{.38\linewidth} \centering \footnotesize{\textit{loop-2d\_s4}, $N_m=6$} \\
  \includegraphics[trim={0 0cm 0 1cm},clip,width=\linewidth]{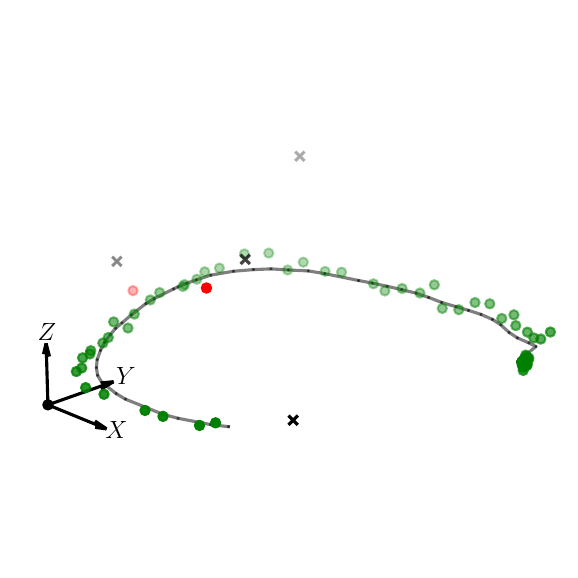}
  \end{minipage}
  \vspace{-1em}
  \begin{minipage}{.22\linewidth}
    \includegraphics[height=3.0cm]{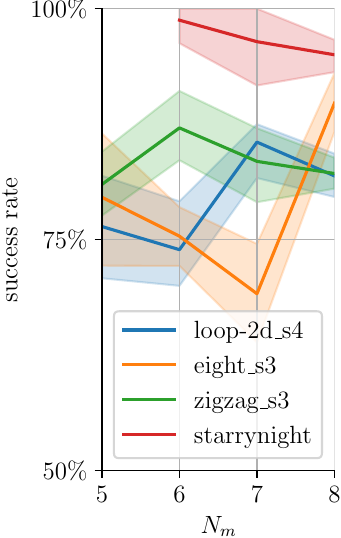}
  \end{minipage}%
  \begin{minipage}{.38\linewidth} \centering \footnotesize{\textit{starrynight}} \\
     \includegraphics[trim={0cm 0 5cm 6cm},clip,height=3.0cm]{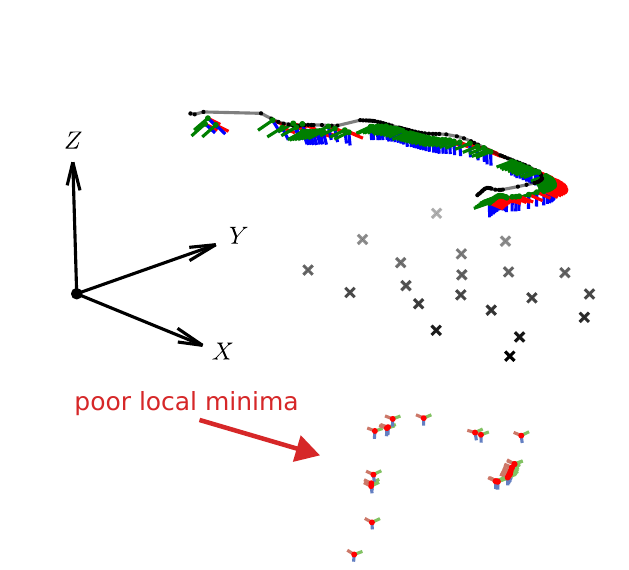}
  \end{minipage}%
  \begin{minipage}{.38\linewidth} \centering \footnotesize{\textit{loop-2d\_s4}}  \\
    \includegraphics[trim={1cm 0 1cm 4cm},clip,height=3.0cm]{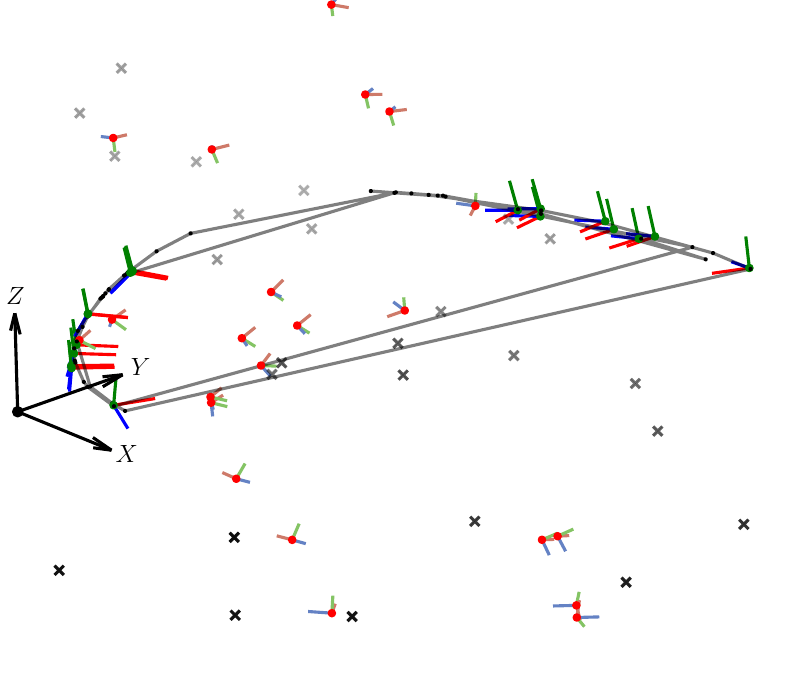}
\end{minipage}
\caption{\pol{Real-world} qualitative study of RO (top row) and stereo localization (bottom row). The left plots show the success rate of the local solver (proportion of convergence to global minima), \pol{as a function of the} number of landmarks $N_m$. For RO, the plots in the middle and right \pol{show} results for the \emph{loop-2d\_s4} dataset, where the success rate depends highly on the number of considered anchors. Global minima are marked with green dots, and (poor, local) minima with red dots. For stereo localization, there is a weaker dependence of the success rate on the number of landmarks, but local minima, shown by small poses with red center, appear less frequently for the \emph{starrynight} dataset, which exhibits less noise. Ground truth trajectories are shown with gray lines, landmarks with black crosses. }\label{fig:real-local-minima}
\end{figure}

We evaluate the performance of the locally optimal \vs~certifiably optimal solver qualitatively in Figure~\ref{fig:real-local-minima}. The local solver has an overall high success rate for the two datasets \emph{eight\_s3} and \emph{zigzag\_s3}, but for \emph{loop-2d\_s4}, it is prone to converge to local minima as the number of anchors drops below $N_m=6$. Intuitively, the cost landscape is more likely to exhibit an approximate symmetry when \pol{all} anchors are close to co-planar, \pol{which is more likely with fewer anchors,} and the local solver gets stuck in the wrong half when initialized there. Note that due to high noise, even the globally optimal position may be far from the ground truth. For stereo localization, the local minima occurring in the \emph{starrynight} dataset are in fact invalid camera poses since the landmarks are not in the estimated field of view. While such faulty estimates could be identified with a manually found heuristic, the local minima occurring in other datasets, for example in \emph{loop-2d\_s4} shown in Figure~\ref{fig:real-local-minima}, are harder to detect with \pol{heuristics.} We also recall that for the latter dataset, the relaxation is not always tight at the present noise levels, and therefore only few estimates can be certified (only the plotted ones). Those few optimal estimates would however be enough to provide an accurate overall trajectory estimate, significantly better than if the poor local minima were used.

Quantitative results are shown in Table~\ref{tab:errors}. We report the average translation and orientation errors, defined as
\newcommand{\errt}{\ensuremath{e_{\vc{t}}}}
\newcommand{\errC}{\ensuremath{e_{\vc{C}}}}
\begin{equation}
  \errt = \|\hat{\vc{t}} - \vc{t}\|_2, \, \errC = \| \hat{\vc{C}}^\top\vc{C} - \vc{I}\|_F,
\label{eq:errors}
\end{equation}
\noindent where $\vc{t},\hat{\vc{t}}$ and $\vc{C},\hat{\vc{C}}$ are the ground truth and estimated translation and orientation, respectively. The average and \ac{STD} are taken over all considered datasets. 
We observe that for both applications, local minima returned by the standard local solver are very poor in accuracy, with errors more than double those of global optima\pol{.}

\begin{table}[b]
  \caption{\pol{Real-world} localization errors of \ac{RO} and stereo localization.  The errors \errt~and \errC~are calculated as in~\eqref{eq:errors}.}\label{tab:errors}
  \adjustbox{max width=\linewidth}{
    \begin{tabular}{lr|r|rr|rr}
& \multicolumn{2}{c|}{RO localization} 
& \multicolumn{4}{c}{stereo localization} \\
\midrule
& global \errt & local \errt 
& global \errt & global \errC & local \errt & local \errC \\
\midrule
\textit{mean} & 0.335296 & 0.945848 &
0.093914 & 0.044693 & 7.676829 & 2.778475 \\
\textit{\ac{STD}} & 0.367695 & 0.506339 &
0.058125 & 0.121845 & 1.681287 & 0.224380 \\
\end{tabular}

  }
\end{table}

To summarize, in both applications, \pol{random} initializations and landmark placements are prone to yield bad, locally optimal solutions. For low-enough noise levels, we can certify globally optimal solutions since our formulation is tight when the automatically identified redundant constraints are used. 

\section{Conclusion and Future Work}\label{sec:discussion}

We have presented new tools to find all possible redundant constraints for a given~\ac{QCQP}, \pol{and applied them to} tighten the semidefinite relaxations of many \pol{state estimation} problems encountered in robotics. The first tool,~\autotight, allows for the fast evaluation of different problem formulations. We have successfully used this tool to evaluate different substitutions for \ac{RO} localization and \pol{found} a novel tight formulation for stereo localization. The second tool,~\autoscale, can be employed to create scalable templates to tighten new setups and larger problem sizes. To show the wide applicability of both tools, we have also evaluated their performance on example problems from the literature~\cite{briales_convex_2017, yang_certifiably_2022}, showing that we find tight relaxations \pol{with} fewer redundant constraints than previously considered. As~\ac{SDP}s scale poorly with the number of constraints, this is an important step to make semidefinite relaxations scale to \pol{problems} encountered in robotics. 

A number of follow-up questions deserve further attention. First, it has been shown that both the measurement graph and the noise level can have an effect on tightness~\cite{cifuentes_local_2022,holmes_efficient_2023,papalia_certifiably_2023}. In future work, we plan to \pol{further} investigate these characteristics using the \pol{new tools, in particular in order} to understand to what level the additional redundant constraints may push the boundaries of tightness. Along the same lines, a given measurement graph may in fact help in finding the variable substitutions and parameters that are most likely to succeed, a component of the proposed method that \pol{currently requires user input.}

Finally, the full potential of the proposed method will be unlocked when faster~\ac{SDP} solvers are developed for problems that require redundant constraints. First steps into this direction have shown promising results~\cite{yang_inexact_2021,yang_certifiably_2022}, but more work remains to be done. In parallel, there lies potential in further pushing the efficiency of  optimality certificates of fast local solvers, for example using sampling-based approaches as in~\cite{cifuentes_sampling_2017} or sparsity-exploiting approaches as in~\cite{holmes_efficient_2023,dumbgen_safe_2023}.

\section*{Acknowledgments}
We would like to thank David Rosen, Alan Papalia, Luca Carlone, Heng Yang, and Nicolas Boumal for helpful discussions around the topic of this paper. We would also like to thank Daniel Frisch for pointing out typos.

\appendix
\subsection{\pol{Additional simulation results}}\label{app:redundant}

\begin{figure}[t]
  \centering
  \includegraphics[width=\linewidth]{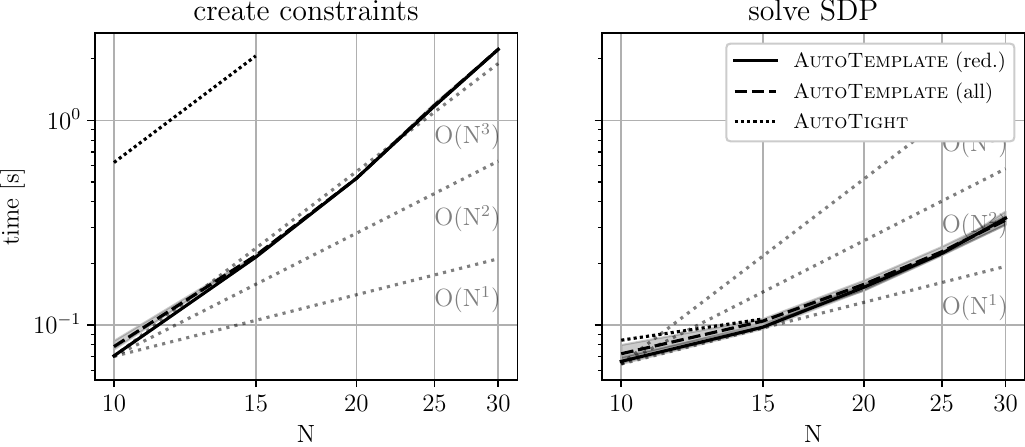}
  \caption{Timing study for~\ac{RO} localization, using the $z_n$ substitution. The same explanations as for Figure~\ref{fig:results-time-ro} apply.}\label{fig:results-time-ro-z}
\end{figure}

\begin{figure}[t]
  \centering
  \includegraphics[width=\linewidth]{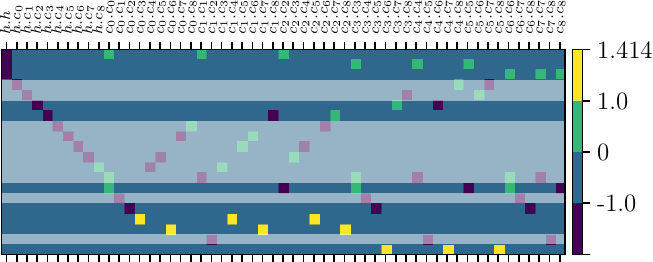}
  \caption{Learned constraint templates for \ppr and \plr~\cite{briales_convex_2017}. The labels $\hom$ and $c_i$ correspond to the homogenization variable and the $i$-th element of $\ve{(\vc{C})}$, respectively. The constraints highlighted in dark are sufficient for rank tightness of \plr, \ppr is rank tight without any redundant constraints.}
  \label{fig:briales-templates}
\end{figure}

\begin{figure}[t]
  \centering
  \includegraphics[width=\linewidth]{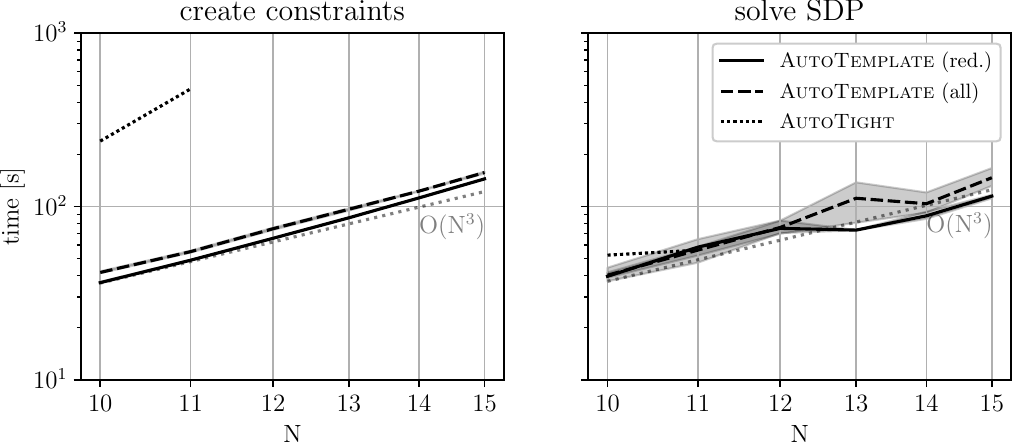}
  \caption{Timing results of scaling to $N$ landmarks for \rplr. Thanks to \autoscale, we can automatically create the constraints of problems up to $N=15$ landmarks.}
  \label{fig:robust-autotemplate-rplr}
\end{figure}

\begin{figure}[t]
  \centering
  \begin{minipage}{.49\linewidth}\centering\footnotesize{\rppr}\end{minipage}
  \begin{minipage}{.49\linewidth}\centering\footnotesize{\rplr}\end{minipage} \\
  \vspace{0.3em}
  \includegraphics[width=.49\linewidth]{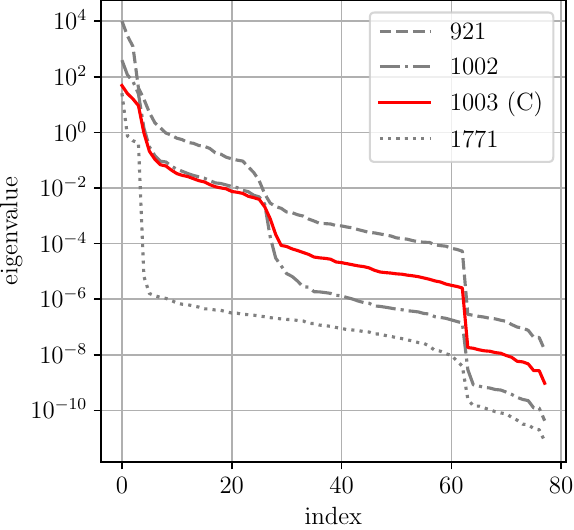}
  \includegraphics[width=.49\linewidth]{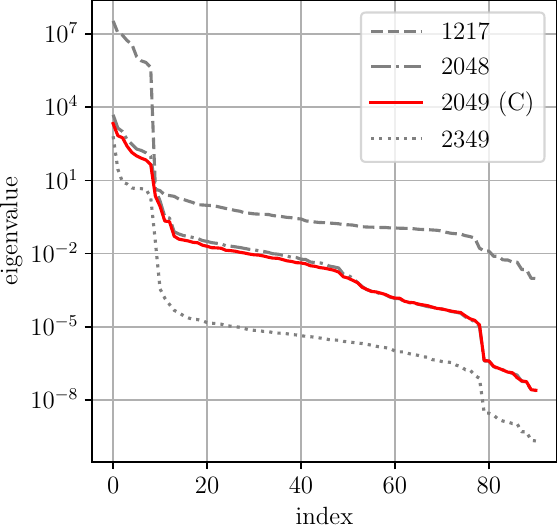}
  \caption{Rank-tightness study for \rppr (left) and \rplr (right). We obtain cost tightness, but not rank tightness, for both problems.}
  \label{fig:robust-eigs}
\end{figure}

Figure~\ref{fig:results-time-ro-z} shows the time required for~\autotight and \autoscale, applied to \ac{RO} localization with $z_n$ substitution. Applying templates and checking for tightness remain relatively cheap as the problem size grows, because the number of total constraints grows only linearly in the number of variables. Even learning the constraints from scratch is reasonably fast for this case. 
Figure~\ref{fig:briales-templates} shows the discovered constraint matrices (in compressed form) for \plr and \ppr. 
Figure~\ref{fig:robust-autotemplate-rplr} shows the timing study for \rplr, and Figure~\ref{fig:robust-eigs} shows the obtained eigenvalue spectra for both \rppr and \rplr. 

\subsection{Mathematical \pol{formulations} of PPR, PLR, rPPR, and rPLR}\label{app:math}

The PPR and PLR problems are derived from~\eqref{eq:briales} by using $\vc{W}_i  = \vc{I}_d$ for \ppr and $\vc{W}_i  = \vc{I}_d - \vc{v}_i\vc{v}_i^\top$ for \plr, where $\vc{v}_i$ is the unit direction of the line. Using $\vc{W}_i = \vc{n}_i\vc{n}_i^\top$, measurements to planes with normal vectors $\vc{n}_i$ could also be modeled with the same framework. The redundant constraints for the relaxation of~\eqref{eq:briales} are given by~\cite{briales_convex_2017}
\begin{subequations}
  \begin{align}
  & \hom^2 = 1 \quad\quad \text{(prim., homogenization),} \\
  & \vc{I}_d = \vc{C}^\top\vc{C} \quad \text{(prim., orthonormal rows),}\label{eq:o1} \\
  & \vc{I}_d = \vc{C}\vc{C}^\top \quad\text{(red., orthonormal columns),} \label{eq:o2} \\
& \pol{\vc{c}_{i}} \times \vc{c}_{i+1|3} = \vc{c}_{i+2|3}, i\inindex{3} \quad \text{(red., handedness),}
 \end{align}
\end{subequations}
where \textit{prim.} and \textit{red.} are short for primary and redundant, $|$ is the modulo operator and $\vc{c}_i$ is the $i$-th column of \vc{C}. This leads to a total of $1 + 2\cdot 6 + 3\cdot3=22$ constraints in 3D, accounting for the symmetry of the optimization variable.

\rppr and \rplr can be written in the form\pol{~\cite{yang_certifiably_2022}}
\begin{equation}
  \min_{\vc{\theta}\in\mathcal{D}} \sum_{i=1}^N \rho\left(r(\vc{\theta},\vc{y}_i)\right),
  \label{eq:robust}
\end{equation}
where $\mathcal{D}$ is the domain of $\vc{\theta}$, $\rho$ is a robust cost function and $r$ the residual function. In order to satisfy the Archimedean condition, the authors further restrict the domain $\mathcal{D}$ to the domain with $\vc{t}\inR{d}$ contained in the ball of radius $T$.\footnote{The Archimedian condition is a stronger form of compactness~\cite{sdp_book}.} For \pol{rPLR}, $\vc{t}$ is additionally chosen so that the landmarks are in the field of view of the camera, characterized by aperture angle $\alpha$. These two problems are thus examples with primary inequality constraints in~\eqref{eq:original}. For~\ac{TLS} cost, it has be shown that solving~\eqref{eq:robust} is equivalent to solving the \ac{QCQP}~\eqref{eq:robust-tls}. This is also true for many other robust cost functions~\cite{yang_certifiably_2022}, and enabled by applying Black-Rangarajan duality~\cite{black_unification_1996}.

\subsection{Proof of Theorem~\ref{thm:1}}\label{app:thm1}

In this proof, We aim to introduce minimal additional notation, and refer the interested reader to~\cite{cifuentes_sampling_2017} for a more in-depth treatment of the following identities. We introduce the variety $\mathcal{V}:=\{\vc{\theta}\,|\, e_i(\vc{\theta}) = 0, i\inindex{N_e}\}$, where $e_i$ are polynomial functions, which corresponds to the feasible set of~\eqref{eq:original}. We define the linear subspace $\mathcal{L}:=\{p(\vc{\theta})\,|\,p(\vc{\theta}) = \vc{\beta}^\top \vech{\vc{x}(\vc{\theta}){\vc{x}(\vc{\theta})}^\top}\}$, where $\vc{\beta}$ is any vector and $\vc{x}(\vc{\theta})$ is the lifted vector as defined in~\eqref{eq:x}, with the dependency on $\vc{\theta}$ made explicit. We will need the property of poisedness, which, loosely speaking, determines whether characteristics derived for samples of a variety hold for any element of it.
\begin{definition}[poisedness]
  Let $\mathcal{R}=\mathbb{R}[\mathcal{V}]$ be the coordinate ring of $\mathcal{V}$, let $\mathcal{L}\subset \mathcal{R}$ be a linear subspace, and let $\vc{\Theta}\subset \mathcal{V}$ be a set of samples. We say that $(\mathcal{L},\vc{\Theta})$ is \textit{poised} if the only polynomial $q\in\mathcal{L}$ such that $q(\vc{\theta}^{(s)})=0, \, \forall \vc{\theta}^{(s)}\in\vc{\Theta}$, is the zero polynomial, \ie, $q(\vc{\theta})=0, \,\forall \vc{\theta}\in\mathcal{V}$.
\end{definition}

As mentioned in Section~\ref{sec:tightness-nullspace}, our sampling method ensures poisedness. For ease of notation, we use $\vc{y}(\vc{\theta}):=\vech{\vc{x}(\vc{\theta}){\vc{x}(\vc{\theta})}^\top}$ in what follows. We create $\vc{Y}$ from the samples in $\vc{\Theta}$ as described in Section~\ref{sec:tightness-nullspace}: $\vc{Y}=\bmat{\vc{y}(\vc{\theta}^{(1)}) & \cdots & \vc{y}(\vc{\theta}^{(S)})}$. The next lemma uses poisedness to ensure that the output of \autotight is guaranteed to encompass all valid constraints.
\begin{lemma}\label{lemma1}
  Let $(\mathcal{L}, \vc{\Theta})$ be poised, and let some polynomial $g(\vc{\theta})\in\mathcal{L}$. Let $\vc{a}$ be such that $g(\vc{\theta})=\vc{a}^\top\vc{y}(\vc{\theta})$. Then we have:
  \vspace{-1em}
  \begin{equation}
    g(\vc{\theta}) = \vc{a}^\top \vc{y}(\vc{\theta}) = 0, \, \forall \vc{\theta} \in \mathcal{V} \iff 
    \vc{a}\in\nullsp{\vc{Y}^\top}.
\end{equation}
\end{lemma}
\noindent
\textit{Proof:} $(\Rightarrow)$ Let $g(\vc{\theta})=0, \, \forall \vc{\theta}\in\mathcal{V}$. Then we also have $g\left(\vc{\theta}^{(s)}\right)=\vc{a}^\top \vc{y}\left(\vc{\theta}^{(s)}\right)=0, \, \forall \vc{\theta}^{(s)}\in\vc{\Theta}$ because all samples are feasible by construction. Since $\vc{Y}=\bmat{\vc{y}\!\left(\vc{\theta}^{(1)}\right) & \cdots & \vc{y}\!\left(\vc{\theta}^{(S)}\right)}$ this implies that $\vc{a}$ is in the nullspace of $\vc{Y}^\top$.

\noindent
$(\Leftarrow)$ Let $\vc{a}\in\nullsp{\vc{Y}^\top}$, or in other words $\vc{a}^\top\vc{Y}=\bmat{g\!\left(\vc{\theta}^{(1)}\right) & \cdots & g\!\left(\vc{\theta}^{(S)}\right)} = \vc{0}^\top$. Since $(\mathcal{L}, \vc{\Theta})$ is poised and $g$ is zero for all samples $\vc{\theta}^{(s)}\in\vc{\Theta}$, we also have $g(\vc{\theta})=0, \, \forall \vc{\theta} \in \mathcal{V}$.$\qed$

Consequently, the elements of the learned matrices by \autotight, $\mathcal{A}_{\ell}$, span the nullspace of $\vc{Y}^\top$. Therefore, any matrix $\vc{B}$ of the set $\mathcal{A}_k$ can be written as a linear combination of the elements of $\mathcal{A}_{\ell}$. This is the main ingredient for proving Theorem~\ref{thm:1}.

\noindent
\textit{Proof of Theorem~\ref{thm:1}:} In Theorem~\ref{thm:1}, (P$_k$) is given by
\begin{equation}
  \begin{aligned}
    \text{(P$_k$)} \quad p_k^\star = \min_{\vc{X}\succeq 0} \, & \langle \vc{Q}, \vc{X} \rangle \\
    \text{s.t. } \langle \vc{B}_j, \vc{X} \rangle &= 0, \langle \vc{A}_0, \vc{X}\rangle=1,
  \end{aligned}
  \label{eq:dualk}
\end{equation}
where $\vc{B}_j$ are the elements of $\mathcal{A}_{k}$ and equivalently, we have 
\begin{equation}
  \begin{aligned}
    \text{(P$_{\ell}$)} \quad p_{\ell}^\star = \min_{\vc{X}\succeq 0} \, & \langle \vc{Q}, \vc{X} \rangle \\
    \text{s.t. } \langle \vc{A}_i, \vc{X} \rangle &= 0, \langle \vc{A}_0, \vc{X}\rangle=1,
  \end{aligned}
  \label{eq:dualell}
\end{equation}
where $\vc{A}_i$ are the elements of $\mathcal{A}_{\ell}$. Calling $q^\star$ the optimal value of~\eqref{eq:qcqp-matrix}, we recall that $p_k^\star\leq q^\star$ and $p_\ell^\star\leq q^\star$, which is true by duality theory, or equivalently because (P$_k$) and (P$_\ell$) are rank relaxations of~\eqref{eq:qcqp-matrix}. Because of Lemma~\ref{lemma1}, (P$_k$) is a relaxation of (P$_\ell$) and we have $p_k^\star \leq p_\ell^\star$. Therefore, if $p_k^\star=q^\star$, then $q^\star=p_k^\star \leq p_{\ell}^\star \leq q^\star$ and all inequalities are equalities. 

Similarly, because (P$_k$) is a relaxation of (P$_\ell$) their respective optimal values respect $\rank{\vc{X}_{k}^\star} \geq \rank{\vc{X}_{\ell}^\star}$. Furthermore, note that the rank of any optimizer $\vc{X}^\star$ of (P$_k$) or (P$_{\ell}$) is at least one, because the homogenization constraint prevents it from being zero. Therefore, if $\rank{\vc{X}_k^\star}=1$ then $1\leq \rank{\vc{X}_{\ell}^\star} \leq \rank{\vc{X}_k}^\star=1$, which concludes the proof.~$\qed$

\bibliographystyle{IEEEtran}
\bibliography{zotero-updated,zotero-final}

\end{document}